\documentclass[journal,twoside,web]{ieeecolor}
\usepackage{tmi}
\usepackage{cite}
\usepackage{amsmath,amssymb,amsfonts}
\usepackage{algorithmic}
\usepackage{graphicx}
\usepackage{textcomp}
\def\BibTeX{{\rm B\kern-.05em{\sc i\kern-.025em b}\kern-.08em
    T\kern-.1667em\lower.7ex\hbox{E}\kern-.125emX}}
\usepackage{diagbox}
\usepackage{graphicx}
\usepackage{multirow}
\usepackage{booktabs}
\usepackage{bm}
\usepackage{amsmath}
\usepackage{amssymb}
\usepackage{wrapfig}
\usepackage{algorithm}
\usepackage{algorithmic}
\usepackage{array} 
\newcommand{\tabincell}[2]{\begin{tabular}{@{}#1@{}}#2\end{tabular}}
\newcommand{\RNum}[1]{\uppercase\expandafter{\romannumeral #1\relax}}
\newtheorem{prop}{Proposition} 
\newtheorem{pro}{Proof} 
\usepackage{subcaption}
\usepackage[table]{xcolor}
\usepackage{tabu}
\usepackage{xcolor}
\usepackage{url}

\label{definition}

\usepackage{xcolor}
\usepackage{xpatch}

\begin{document}
\title{MedRDF: A Robust and Retrain-Less Diagnostic Framework for Medical Pretrained Models Against Adversarial Attack}
\author{Mengting Xu, Tao Zhang, and Daoqiang Zhang
\thanks{This work was supported by the National Natural Science Foundation of China (Nos. 62136004, 61876082, 61732006), the National Key R\&D Program of China (Grant Nos. 2018YFC2001600, 2018YFC2001602), and also by the CAAI-Huawei MindSpore Open Fund.}
\thanks{Mengting Xu, Tao Zhang, and Daoqiang Zhang are with the College of Computer Science and Technology, Nanjing University of Aeronautics and Astronautics, Nanjing 211106, China (e-mail: \{xumengting, dqzhang\}@nuaa.edu.cn).}
%\thanks{S. B. Author, Jr., was with Rice University, Houston, TX 77005 USA.
%He is now with the Department of Physics, Colorado State University,
%Fort Collins, CO 80523 USA (e-mail: author@lamar.colostate.edu).}
%\thanks{T. C. Author is with the Electrical Engineering Department,
%University of Colorado, Boulder, CO 80309 USA, on leave from the National
%Research Institute for Metals, Tsukuba, Japan (e-mail: author@nrim.go.jp).}
\thanks{Daoqiang Zhang is the corresponding author.}
}

\maketitle

\begin{abstract}
%Deep neural networks  have been widely used in challenging tasks such as computer-aided disease diagnosis based on medical images.
%However, deep neural networks are discovered to be non-robust when attacked by imperceptible adversarial examples, which is dangerous for medical diagnostic system that requires high reliability.
%Accordingly, it is necessary to design a highly robust and reliable framework for medical diagnostic tasks.
%In this paper, we propose a \textbf{R}obust \textbf{D}iagnostic \textbf{F}ramework for \textbf{Med}ical image against adversarial attack (i.e., MedRDF). 
%Specifically, 
%MedRDF firstly creates pre-processed copies of the medical image, and obtains the output labels of these copies from the pre-trained medical diagnostic model. Then, MedRDF outputs the final robust diagnostic result of the medical image based on the labels of these copies by majority voting.
%What's more, in addition to the diagnostic result, MedRDF produces the Robust Metric (RM) as the confidence of the result.
%Therefore, it is convenient and reliable to utlize MedRDF to convert pre-trained non-robust diagnostic models into robust ones.
%The experimental results on COVID-19 and DermaMNIST datasets verify the effectiveness of our MedRDF in improving the robustness of medical diagnostic models.

Deep neural networks are discovered to be non-robust when attacked by imperceptible adversarial examples, which is dangerous for it applied into medical diagnostic system that requires high reliability.
However, the defense methods that have good effect in natural images may not be suitable for medical diagnostic tasks. The pre-processing methods (e.g., random resizing, compression) may lead to the loss of the small lesions feature in the medical image. Retraining the network on the augmented data set is also not practical for medical models that have already been deployed online.
Accordingly, it is necessary to design an easy-to-deploy and effective defense framework for medical diagnostic tasks.
In this paper, we propose a \textbf{R}obust and \textbf{R}etrain-Less \textbf{D}iagnostic \textbf{F}ramework for \textbf{Med}ical pretrained models against adversarial attack (i.e., MedRDF). It acts on the inference time of the pertained medical model.
Specifically, for each test image, 
MedRDF firstly creates a large number of noisy copies of it, and obtains the output labels of these copies from the pretrained medical diagnostic model. Then, based on the labels of these copies, MedRDF outputs the final robust diagnostic result by majority voting.
In addition to the diagnostic result, MedRDF produces the Robust Metric (RM) as the confidence of the result.
Therefore, it is convenient and reliable to utilize MedRDF to convert pre-trained non-robust diagnostic models into robust ones.
The experimental results on COVID-19 and DermaMNIST datasets verify the effectiveness of our MedRDF in improving the robustness of medical diagnostic models.
\end{abstract}


\section{Introduction}
\label{sec:introduction}
\begin{figure*}[t]
	\includegraphics[width=\textwidth]{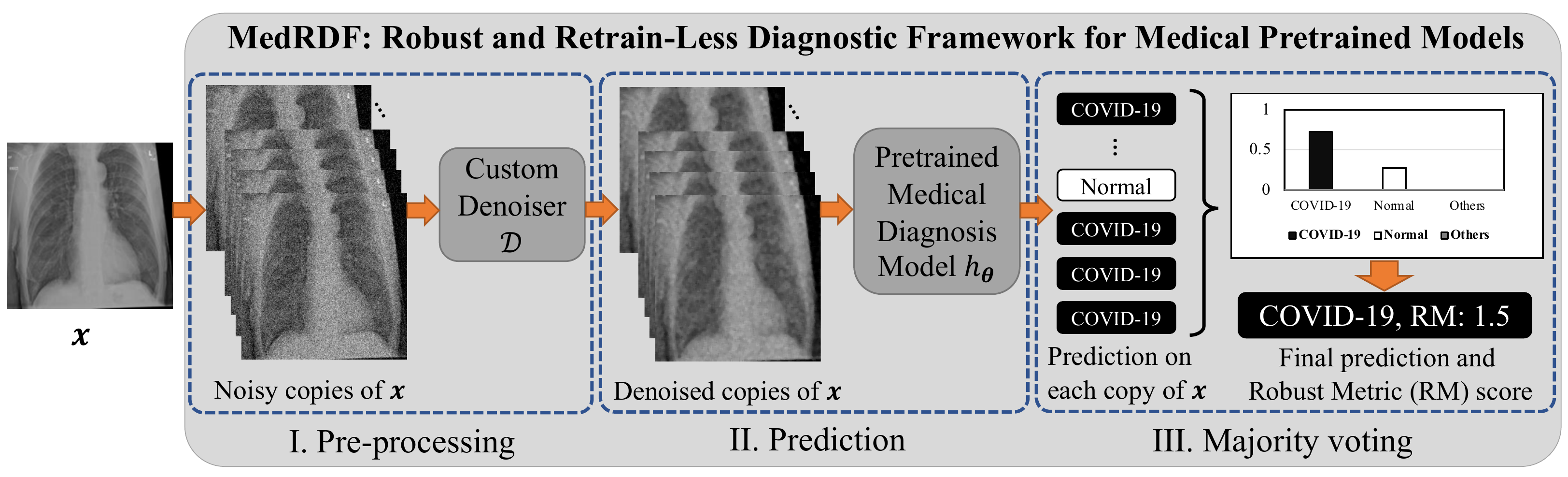}
	\caption{The Robust and Retrain-Less Diagnostic Framework for Medical pretrained models (MedRDF). 
		\RNum{1}, each test medical image $\bm{x}$ is perturbed by isotropic noises $\bm{\eta}$ to produce the noisy copies of $\bm{x}$, then they are denoised by the pre-defined denoiser $\mathcal{D}$. \RNum{2}, the denoised copies are input to the pre-trained medical diagnostic model $h_{\bm{\theta}}$ to get the predictions. \RNum{3}, the robust diagnostic result $g(\bm{x})$ on $\bm{x}$ and the Robust Metric (RM) of the result are obtained by the majority voting on the prediction labels of denoised ones.	
	} 
	\label{fig:structure}
\end{figure*}
\IEEEPARstart{T}{here} are many impressive examples of deep neural networks achieving excellent performances on medical diagnostic tasks in  radiology \cite{rajpurkar2017chexnet}, dermatology \cite{esteva2017dermatologist}, and ophthalmology \cite{gulshan2016development}, etc. 
However, recent studies have revealed the fact that the robustness of the state-of-the-art neural network is poor, i.e., it is easily to craft a visually imperceptible adversarial example to mislead a well-trained network with high confidence~\cite{szegedy2014intriguing,madry2018towards,carlini2017towards}.
The vulnerability to adversarial examples poses a huge threat to the deployment of these models to medical diagnostic tasks that require extremely high reliability~\cite{paschali2018generalizability,xu2021towards,ozbulak2019impact,li2020defending}. 
For example, the misdiagnosis of COVID-19 may cause a large number of diseases to spread.
Therefore, developing a robust model to defend against adversarial attacks is very crucial in medical image field.

There are many different defense strategies developed in natural image field. One of the most successful empirical defenses to date is adversarial training~\cite{madry2018towards}, which can be regarded as a data augmentation technique that trains neural networks on adversarial examples. 
However, adversarial training in medical image is problematic as it requires a large labeled training set whereas medical data sets are usually with a small amount of labeled samples. 
To solve this problem, Li et al. \cite{li2020defending} propose the semi-supervised adversarial training (i.e., SSAT) which utilizes both labeled and unlabeled data to generate psudo-labels. 
However, the application of SSAT is also limited, because for most medical diagnostic tasks, unlabeled data is also inaccessible, not to mention the heterogeneity between multi-site data sets acquired through different devices (i.e., data distribution difference) and the privacy of medical data. 
Moreover, Xue et al. \cite{xue2019improving}  propose a defense mechanism which embeds an auto-encoder into the model structure and keeps high-level features invariant to general noises.
However, retraining mechanism is not friendly to the medical diagnostic model that has been already deployed online. It is time-consuming and laborious to go back to the online process. 
Other pre-processing based-defense methods have also shown effectiveness in natural image field. For example, Xie et al.~\cite{xie2018mitigating} use random resizing and padding (Random R-P) to pre-process the input images before feeding the images into the models. Jia et al. propose the ComDefend~\cite{jia2019comdefend}  to transform the adversarial image to its clean version by compression and reconstruction. However, these defense methods that have good effects in the field of natural images may not be suitable for medical images. 
For natural images,  there is strong similarity and relevance between neighbor pixels in the local structure, random resizing and image compression can help reduce the redundant information of the image, while retaining the dominant information.
But for medical images, medical lesions often occupy only a few pixels. Random resizing and image compression may cause the loss of lesion features, thereby affecting the classification and defense effects. 
To make matters worse, there is still no effective confidence indicator for doctor to evaluate the diagnostic result of the model.
Therefore, how to reliabily improve the robustness of medical diagnostic model is still an open problem.

In this paper, we propose a novel Robust and Retrain-Less Diagnostic Framework for Medical Pretrained Models (i.e., MedRDF) to defense against adversarial attack. 
As shown in Fig.~\ref{fig:structure}, our proposed MedRDF can easily convert the non-robust pre-trained model to robust one in inference time without retraining. 
Specifically, \textit{firstly}, for each queried medical image $\bm{x}$, 
MedRDF produces a large number of copies (i.e., with adding common noise and denoising) around it. \textit{Secondly}, the denoised copies are input into the pre-trained diagnostic model to get the prediction labels. \textit{Finally}, MedRDF outputs the robust diagnostic result of medical image $\bm{x}$ by majority voting on the prediction labels of denoised ones.
What's more, MedRDF also produces the Robust Metric (RM) as the confidence of the result, which can be used to instruct the doctor to adopt the diagnostic result or re-evaluate it.

The main innovations of our MedRDF can be summarized as follows:
\begin{itemize}
	\item A novel Robust and Retrain-Less Diagnostic Framework for medical pretrained models (i.e., MedRDF) has been proposed. The MedRDF can be applied to all medical diagnostic tasks seamlessly without retraining diagnostic models, which is very convenient for diagnostic services that are already deployed online.
	\item A novel Robust Metric (i.e., RM) based on MedRDF has been proposed. It can give the confidence score of the diagnostic result produced by MedRDF, so as to guide the following work of the doctor, such as adopting the result (with high RM) or re-evaluating this case (with low RM).
\end{itemize}

%The rest of the paper is organized as follows: Section~\ref{sec: related work} introduces the related works; Section~\ref{sec:materials} describes the studied materials.

%Section III describes the studied materials and our proposed DA-MIDL method; Section IV shows the experimental settings and results for multiple AD diagnosis tasks compared with several state-of-the-art meth- ods; Section V presents the discussion on the effectiveness of our attention modules, identified pathological locations and limitations; Section VI concludes the work.

\section{Related work}
\label{sec: related work}
In this section, we first briefly introduce the latest developments in deep learning in the diagnosis of coronavirus disease (COVID-19) and common pigmented skin lesions. Then, the recent adversarial attacks and defense methods on natural and medical images have been reviewed.

\subsection{Deep Learning for Medical Image Analysis}
In the past few years, high-performance deep diagnostic classification models for disease diagnosis have emerged. Here, we are going to introduce two successful applications of deep learning models in medical image analysis.
\subsubsection{Coronavirus Disease (COVID-19)}
In recent years, the global outbreak of the Coronavirus disease (COVID-19) has caused tens of thousands of deaths and infected millions of people around the world. This undoubtedly poses a huge threat to the lives of the human beings and the national public health system. Any technical tool that can quickly screen for COVID-19 infection with high accuracy is vital to healthcare professionals. The main clinical tool currently used to diagnose COVID-19 is Reverse Transcription Polymerase Chain Reaction (RT-PCR), but it is expensive, less sensitive, and requires specialized medical personnel~\cite{chowdhury2020can}. A clinical study of COVID-19 infected patients showed that most of these types of patients were infected by lung infections after being exposed to the virus\cite{jain2021deep}. Therefore, easy-to-use and low-cost X-ray (i.e., radiography) imaging has become an excellent alternative to COVID-19 diagnosis.

Many automatic algorithms have been proposed to diagnose COVID-19 from chest X-ray images~\cite{apostolopoulos2020covid,ozturk2020automated,ucar2020covidiagnosis}. In particular, deep learning methods have been considered the best performing methods~\cite{jamshidi2020artificial}, including Generative Adversarial Networks (GANs)~\cite{creswell2018generative}, Extreme Learning Machine (ELM)~\cite{huang2006extreme}, and Long /Short Term Memory (LSTM)~\cite{hochreiter1997long}.
Besides,  Jain et al.~\cite{jain2021deep} compared Inception V3~\cite{szegedy2016rethinking}, Xception~\cite{chollet2017xception}, and ResNeXt~\cite{xie2017aggregated} models which have high performance in natural image field and
examined their accuracy in diagnosis of COVID-19.
Morever, Schlemper et al.~\cite{schlemper2018attention} proposed the Attention-Gated Sononet (AG-Sononet) model, which is carefully designed for fetal ultrasound images. It can also be used for COVID-19 disease diagnosis.

\subsubsection{Pigmented Skin Lesions}
Skin cancer is one of the most commonly diagnosed cancers worldwide. 
According to the 2019 statistical report of the American Association, the number of new cases and deaths of skin cancer in the United States (excluding basal cell and squamous cell skin cancer) is as high as 104,350 and 11,650, respectively~\cite{siegel2019cancer}. Among them, melanoma accounts for the largest proportion of all lesions, and the estimated number of new cases and deaths are 92.5\% and 62.1\%, respectively. However, the skin cancer can be highly treated by early detection and diagnosis, thus reducing the mortality rate.

Due to the importance of early detection, many deep learning methods are used to improve the accuracy of diagnosis and expand the scale of diagnosis. 
%In 2016 and 2017, the International Skin Imaging Collaboration (ISIC) hosted the first public benchmark for melanoma detection in dermoscopic images titled "Skin Lesion Analysis Towards Melanoma Detection" ~\cite{gutman2016skin,codella2018skin}, which attracted over 900 registrations worldwide, and more than 350 submissions. In 2018, in the challenge hosted by Medical Image Computing and Computer Aided Intervention (MICCAI) conference in Granada, Spaini, in addition to the increasing in the size of the dataset and number of diagnostic labels, key changes to evaluation criteria and study design were implemented to better reflect the complexity of clinical scenarios encountered in practice. 
For example, Li et al.~\cite{li2018skin} proposed a framework consisting of
multi-scale fully-convolutional residual networks and a lesion index calculation unit (LICU)
to simultaneously address lesion segmentation and lesion classification.
Yan et al.~\cite{yan2019melanoma} proposed an attention-based melanoma recognition method, which introduces an end-to-end trainable attention module regularization for melanoma recognition.

\subsection{Adversarial Attack}
Despite the high performance of deep neural networks in medical image diagnosis,
Szegedy et al.~\cite{szegedy2014intriguing} first discovered that deep networks are extremely vulnerable to the adversarial examples. The so-called ``adversarial example" is added carefully designed perturbation on the original example, which is invisible to the human eye, thus misleading the network output a wrong perdiction with a high confidence. Even worse, due to the transferability of adversarial examples, the perturbation designed for one network can also be used to fool other networks.

In recent years, the adversarial attack methods for natural image have developed rapidly. 
Goodfellow et al.~\cite{goodfellow2014explaining} proposed Fast Gradient Sign Method (FGSM) to generate adversarial examples.
It calculates the gradient of the loss function with respect to the pixel, and modifies the pixel value of a fixed step along the direction of the gradient. Based on this work, Madry et al.~\cite{madry2018towards} proposed an iterative attack method, which randomly starts a perturbation, and updates the pixel value multi-time along the direction of the gradient, which is called the Projected Gradient Descent (PGD). In addition to these gradient-based methods, Carlini and Wagner (C\&W attack)~\cite{carlini2017towards} explored the use of maximum marginal loss and optimization method to generate adversarial examples with high fooling rate and small distortion with respect to the original image.
In addition, more and more black-box attacks have been proposed. These so-called black-box attack can successfully change the model prediction without knowing the parameters and structure of the attacked model.
Uesato et al.~\cite{uesato2018adversarial} proposed simultaneous perturbation stochastic approximation (SPSA) attack. It is a gradient-free query-based attack method, which minimizes the output logits of the true label and the largest logits of the rest of labels. 
Chen et al.~\cite{chen2020rays} proposed the hard-label RayS attack, which only relies on the hard-label output of the target model and utilizes a fast check step to skip unnecessary searches. This significantly saves the number of queries needed for the hard-label attack. 

Apart from the development of adversarial attack in the field of natural images, medical image domian has also payed more and more attention to this topic. 
Ma et al.~\cite{ma2021understanding} analyzed the different behaviors of medical images and natural images when attacked by adversarial examples, and concluded that medical images are more vulnerable to adversarial attacks.
Other studies~\cite{paschali2018generalizability,taghanaki2018vulnerability,xu2021towards} evaluated the robustness of deep diagnostic models on different tasks by adversarial attacks.

\subsection{Adversarial Defense}
Considering the importance of network robustness, many defense methods have been proposed~\cite{papernot2016distillation,buckman2018thermometer,shafahi2019adversarial}. Among which, Adversarial Traing (AT) has been demonstrated to be one of the most effective defense methods. AT can be regarded as a data augmentation technology, that trains network on adversarial examples. 
After that, many methods were improved based on AT and showed superior performance. TRADES~\cite{zhang2019theoretically} trades adversarial robustness off against accuracy. The objective function of TRADES is a linear combination of natural loss and regularization term.
MART~\cite{wang2019improving} differentiates the misclassified examples and correctly classified examples during adversarial training and adopts a regularized adversarial loss involving both adversarial and natural examples to improve the robustness of models.
For medical image field, Liu et al.~\cite{liu2020no} propose the augmentation method to add adversarial synthetic nodules and adversarial attack samples to the training data to improve the generalization and the robustness of the lung nodule detection systems.
However, these methods require retraining the model, which is not friendly to the medical diagnostic models that have been already deployed online.

Besides adversarial training methods, many pre-processing based-defense methods have been proposed. Xie et al.~\cite{xie2018mitigating} use random resizing and padding (Random R-P) to pre-process the input images before feeding the images into the models to make predictions. Jia et al. propose the ComDefend~\cite{jia2019comdefend}, which consists of a compression convolutional neural network (ComCNN) and a reconstruction convolutional neural network (RecCNN) to transform the adversarial image to its clean version. 
However, the random resizing and compression operators may loss the lesion features of medical images.

\section{Materials}\label{sec:materials}
In this section, we will introduce in detail the datasets and pre-trained models used in our study.
\subsection{Datasets}
Two public datasets are used in this study, including:
\subsubsection{COVID-19 Radiography Database \cite{chowdhury2020can}}
It consists of  chest X-ray images with size $224 \times 224$ of COVID-19 positive, normal, and viral pneumonia images (i.e., 3-class diagnostic task). 
%The COVID-19 cases, are created from the following four major data sources: 1) Italian society of medical and interventional radiology (SIRM) COVID-19 database~\cite{SIRM}, 2) novel corona virus 2019 dataset~\cite{COVID-Chestxray}, 3) COVID-19 positive chest X-ray images from different articles~\cite{COVID-19ChestX-RayDatabase}, and 4) COVID-19 chest imaging at thread reader []. The normal cases and viral pneumonia cases are obtained from RSNA-Pneumonia-Detection-Challenge [] and Kaggle chest X-ray database [].
In the current release, there are $1200$ COVID-19 positive images, $1341$ normal images, and $1345$ viral pneumonia images. We have split this dataset into training, validation, and test set with ratio $8:1:1$. 
%We report classification accuracy of the original and adversarial test dataset to validate the effectiveness of our proposed MedRDF.
\subsubsection{DermaMNIST~\cite{yang2021medmnist}}
It is based on HAM10000~\cite{tschandl2018ham10000,codella2019skin}, 
which consists of 10, 015 multi-source dermatoscopic images of common pigmented skin lesions. 
This dataset is labeled as 7 different categories (i.e., actinic keratoses, basal cell carcinoma, benign keratosis, dermatofibroma, melanocytic nevi, melanoma, vascular), as a 7-class classification task. 
We have split the images into training, validation and test set with ratio 7 : 1 : 2. The source images of 3 × 600 × 450 are resized into 3 × 28 × 28.

\subsection{Pre-trained Models}
In order to better explore the effect of MedRDF on different pretrained models,
the base classifiers we use in experiments are natural image based ResNet-18 and ResNet-50 \cite{he2016deep} and medical image based AG-Sononet-16 \cite{schlemper2018attention}.
We directly train the networks on COVID-19 and DermaMNIST datasets without fine-tuning. The ResNet-18 and AG-Sononet-16 are trained for 100 epochs using stochastic gradient descent with momentum 0.9 and weight decay $1e^{-6}$. The initial learning rate is $1e^{-4}$ and is decayed by 0.1 on 50 and 75 epochs. 
The ResNet-50 is trained for 100 epochs using stochastic gradient descent with momentum 0.9 and weight decay $1e^{-4}$. The initial learning rate is $1e^{-3}$ and is decayed by 0.1 on 50 and 75 epochs. The batch size is 10.
%These models are trained with the SGD optimizer with learning rate $1e^{-4}$, the batch size is 10, and the total epoch is 100. 
%The initial learning rate is 0.01 and is decayed by 0.1 every 10 epochs. 

%\subsection{Attack methods}

\section{Methods}
\label{methods}
\subsection{Problem Formulations}
Let $\mathcal{X} \in \mathbb{R}^d$ denote the input space and $\mathcal{Y}=\{1,\cdots, K\}$ be a finite set consists of $K$ possible class labels. $D=\{(\bm{x}_1,y_1),\cdots,(\bm{x}_m,y_m)\}$ is a training set with $m$ labeled examples, where $\bm{x}_i\in \mathcal{X}$ is the feature vector and $y_i\in \mathcal{Y}$ is the label of the $i$-th example.  Given a medical diagnostic model $h_{\bm{\theta}}$ with parameters $\bm{\theta}$, it outputs the class label $h_{\bm{\theta}} (\bm{x}_i)$ for each input image $\bm{x}_i\in \mathcal{X}$:
\begin{equation}
h_{\bm{\theta}}(\bm{x}_i)=\mathop {\arg\max}_{k=1,\cdots,K} \bm{p}_k(\bm{x}_i,\bm{\theta}),
\end{equation}
%\begin{equation}
%\bm{p}_k(\bm{x}_i,\bm{\theta})=\exp(\bm{z}_k(\bm{x}_i,\bm{\theta}))/ \sum_{k^\prime=1}^K \exp(\bm{z}_{k^\prime}(\bm{x}_i,\bm{\theta})),
%\end{equation}
%where $\bm{z}_k(\bm{x}_i,\bm{\theta})$ is the logits output of the network with respect to class $k$, and 
where $\bm{p}_k(\bm{x}_i,\bm{\theta})$ is the probability (softmax on logits) of $\bm{x}_i$ belonging to class $k$. We denote $\mathcal{A}_{h_{\bm{\theta}}}$ as the space of adversarial examples for the pre-trained model $h_{\bm{\theta}}$. Adversarial example $\bm{x}^\prime \in \mathcal{A}_{h_{\bm{\theta}}}$ is supposed to be quasi-imperceptible to the human eye and misclassified by $h_{\bm{\theta}}$, i.e.,
\begin{equation}
d(\bm{x}, \bm{x}^\prime)\le \epsilon \quad \text{and} \quad h_{\bm{\theta}} (\bm{x}) \ne h_{\bm{\theta}} (\bm{x}^\prime),
\end{equation}
where $d(\cdot)$ is the distance function, $\epsilon$ is the maximum perturbation for adversarial attack. Here we aim to design a robust diagnostic framework $g$ to correctly classify these adversarial examples $\bm{x}^\prime \in \mathcal{A}_{h_{\bm{\theta}}}$ with the pre-trained model $h_{\bm{\theta}}$.

\subsection{Framework Details}
Inspired by random smoothing \cite{cohen2019certified}, 
we construct a robust and retrain-less diagnostic framework (MedRDF) $g$ for medical pretrained model $h_{\bm{\theta}}$. Fig. \ref{fig:structure} illustrates the flowchart of proposed MedRDF $g$.
Specifically, MedRDF returns the class label which the base classifier $h_{\bm{\theta}}$ is most likely to return. \textit{Firstly}. MedRDF perturbs input $\bm{x}$ by isotropic noise $\bm{\eta}$ from distribution $\mu$ to produce the noisy copies, and then denoises the noisy copies by pre-defined denoiser $\mathcal{D}$. \textit{Secondly}. MedRDF inputs these copies to pre-trained medical diagnostic model $h_{\bm{\theta}}$ to obtain the prediction labels. \textit{Thirdly}, the final diagnostic result is obtained by majority voting based on the labels of denoised copies.
The MedRDF $g$ is formulated as follows:
\begin{equation}
\begin{split}
g(\bm{x})=\arg\max_{k \in \mathcal{Y}} &\mathbb{P}(h'_{\bm{\theta}}(\bm{x} + \bm{\eta}) = k),  \\
\bm{\eta} & \sim \mu(0,\sigma \bm{I}),
\end{split}
\label{equ:g(x)}
\end{equation}
where $h'_{\bm{\theta}}(\cdot)$ represents $h_{\bm{\theta}}(\mathcal{D}(\cdot))$, $\mathcal{D}(\cdot)$ is the pre-defined denoiser.
An equivalent definition is that $g(\bm{x})$ returns the class $k$ whose pre-image $\{\bm{x}+\bm{\eta} \in \mathbb{R}^d: h'_{\bm{\theta}}(\bm{x}+\bm{\eta}) = k\}$ has the largest probability measure under the distribution $\mu(\bm{x},\sigma \bm{I})$.
The level of noise $\bm{\eta}$ has been bounded by $\sigma$, where the noise level $\sigma$ controls the tradeoff between robustness and accuracy, i.e., the robustness of the MedRDF increases with the increasing of $\sigma$ while its standard accuracy decrease. 

The detailed information of our MedRDF is described as follows:

\subsubsection{Isotropic Noise $\eta$} Recent studies \cite{cao2017mitigating,szegedy2014intriguing} show that the non-robustness of deep networks against attacks is caused by the high nonlinearity of deep networks. 
\begin{figure}[t]%靠文字内容的左侧
	\centering
	\includegraphics[width=0.4\textwidth]{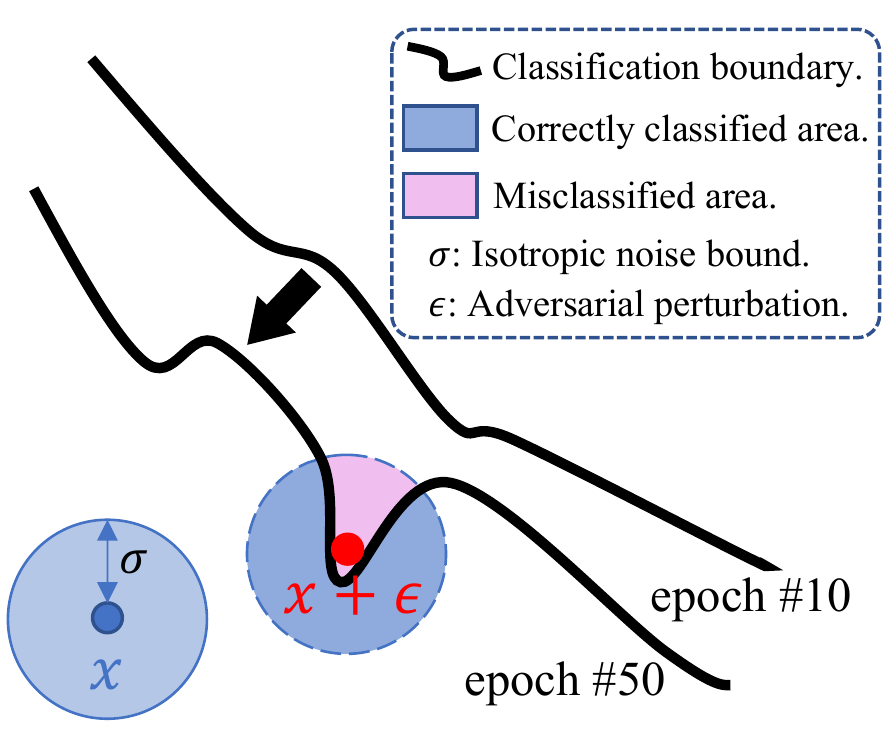}
	\caption{The effect of isotropic noise. A significant increase in the curvature of the decision boundary during continuous training from epoch $\#10$ to epoch $\#50$. The hypercubes centered at $\bm{x}+\bm{\epsilon}$ and $\bm{x}$ intersect the most with the area that examples can be correctly classified.}
	\label{fig:noise}
\end{figure}
\begin{algorithm} [t]
	\caption{MedRDF: Robust and Retrain-Less Diagnostic Framework for Medical pretrained models}  
	\label{alg:framework}
	\begin{algorithmic}
		\STATE {\bfseries Input} {
			base classifier $h_{\bm{\theta}}$, diagnostic case $\bm{x}$, noise distribution $\mu(0,\sigma \bm{I})$, sampling numbers $n$, abstention threshold $\alpha$, denoiser operator $\mathcal{D}$.
		}
		\STATE 	Initialization array: $\text{counts}[0, \cdots, n-1]$
		\FOR{$i\leq n$}
		\STATE Sample noise $\eta_i \sim \mu(0, \sigma \bm{I})$
		\STATE $\text{counts}[h_{\bm{\theta}} (\mathcal{D}(\bm{x}+\eta_i))]++$
		\ENDFOR
		\STATE $n_A, n_B \leftarrow $ top two indices in counts
		\STATE \textbf{if } {$\text{Binom}(n_A, n_A + n_B, 1/2) \le \alpha $ \textbf{Output} $k_A$ }
		\STATE \textbf{else return -1 (ABSTAIN)}
	\end{algorithmic}
\end{algorithm}
\begin{table}[h]
	\centering
	\caption{Classification accuracy (\%) of the MedRDF $g$ on COVID-19 with salt-and-pepper noise (with level as $\sigma$) and median filter.  The attacker is crafted by PGD with $100$ steps and maximum $L_\infty$ perturbation $\epsilon$ on ResNet-50. } 
	\centering  
	{
		\begin{tabular}{c|c|c|c|c|c|c}
			\toprule
			\diagbox{$\epsilon$}{$\sigma$}& $0.05$ & $0.1$ & $0.15$ & $ 0.2$  & $0.25$ & $0.3$ \\
			\midrule
			$0 $ & \textbf{93.4} & 91.6& 77.4& 50.0&45.8 & 30.0 \\
			$2/255$ & \textbf{93.0} & 91.8 & 79.4 & 51.6 & 45.8 & 44.2 \\
			$4/255$ & \textbf{92.4} & 91.8  & 82.0 & 54.0 & 46.0 & 44.0 \\
			$8/255$ & 91.1 & \textbf{91.2} & 84.8 & 59.6 & 47.0 & 44.2 \\
			$16/255$ & 87.8 & \textbf{91.4} & 89.4 & 73.0 & 51.2 & 46.6 \\
			\bottomrule
		\end{tabular}
	}
	\label{table:dnoise}
\end{table}

\begin{table}[h]
	\centering
	\caption{Accuracy (\%) and test time (s) on each image of MedRDF on different number of copies. The base classifier is ResNet-18. The common noise is salt-and-pepper noise with $\sigma=0.1$, the denoiser in median filter, and maximum $L_\infty$ perturbation $\epsilon=8/255$.  The number after attack method represents the number of iteration steps. The bold number represents the result of our selection.} 
	\centering  
%	\resizebox{\textwidth/2}{16mm}
	\setlength{\tabcolsep}{1.5mm}
	{
		\begin{tabular}{c|c|c|c|c|c|c}
			\toprule
			Datasets& $n$ & Natural & I-FGSM-7 & PGD-7 & C\&W & Time(s)\\
			\midrule
			\multirow{4}{*}{COVID-19}& $1e^2$& 91.2 & 85.4& 92.2& 89.2&0.1\\
			&$1e^3$&91.2& 86.2& 93.0&90.0&1.1\\
			&\textbf{$1e^4$}&\textbf{91.2}& \textbf{86.2}& \textbf{93.2}&\textbf{90.4}&\textbf{3.8}  \\
			&$1e^5$&91.4& 86.2& 93.2&90.0&87.6 \\
			\midrule
			\multirow{4}{*}{DermaMNIST}& $1e^2$& 68.9 & 61.3& 65.4& 64.3&0.1\\
			&$1e^3$&69.0& 62.5& 67.0&65.6&0.1 \\
			&\textbf{$1e^4$}&\textbf{69.0}& \textbf{63.1}& \textbf{67.3}&\textbf{66.0}& \textbf{1.1} \\
			&$1e^5$&70.4& 63.2& 67.3&65.9&10.0 \\
			\bottomrule
		\end{tabular}
	}
	\label{table:n}
\end{table}
Kalimeris et al. \cite{kalimeris2019sgd} show that with the continuous training of the network, a significant increasing in the curvature of the decision boundary and loss landscape will occur, and the adversarial examples are easy to hide in these isolated regions with high curvature \cite{moosavi2019robustness}, as illustrated in Fig. \ref{fig:noise}. 
Based on this observation, we add the common random noise $\eta$ bounded by $\sigma$ to original image, which can reduce the impact of the adversarial example in isolated area on the accuracy of the model.
%, during the inference process of medical diagnostic model.
As shown in Fig. \ref{fig:noise}, in the noise area, with $\bm{x}$ and $ \bm{x} + \epsilon$ as the center and maximum noise $\sigma$ as the boundary, most examples can be correctly classified. The result is also true for the adversarial example in the isolated area.
Therefore, adding isotropic noise to the original image to generate noisy copies can effectively
instruct the network not to be misled by adversarial examples.
However, although neural networks have certain robustness to common noise, too large noise will still lead to the accuracy decrease of $h_{\bm{\theta}}$, which will also affect the final prediction result of $g$ based on $h_{\bm{\theta}}$. In the following subsection, we will introduce the denoising operator to alleviate the decline of accuracy.

\subsubsection{Pre-defined Denoiser}  To alleviate the accuracy decline of the base classifier $h_{\bm{\theta}}$ under large isotropic noise, denoising operator has been adopted in our MedRDF. 
Instead of CNN-based denoiser \cite{salman2020denoised}, we use Gaussian Smoothing (GS) and median Filter (MF) as denoisers in our work, which have faster inference speed and more efficient GPU memory than CNN-based denoiser. 
%We adopt $h'_{\bm{\theta}}(\cdot)$ to represent $h_{\bm{\theta}}$ with denoiser operator, i.e., $h_{\bm{\theta}}(\mathcal{D}(\cdot))$. 

\subsubsection{Prediction and Majority Voting} 
For notational convenience, we define Equation~(\ref{equ:g(x)}) as $P_k = \mathbb{P}(h'_{\bm{\theta}}(\bm{x} + \bm{\eta}) = k)$. Let $\hat{k_A} = \arg \max_k P_k$. Notice that by definition, $g(x) = \hat{k_A}$.
We draw $n$ noise examples with Markov Chain Monte Carlo (MCMC) principle from distribution $\mu(0,\sigma {\bm{I}})$, and inquire $n$ noise-corrupted copies of $\bm{x}$ through the base classifier $h'_{\bm{\theta}}(\cdot)$. 
Sample a vector of class counts $\{n_k\}_{k\in \mathcal{Y}}$ from Multinomial$(\{P_k\}_{k\in \mathcal{Y}}, n)$.
Let $k_A = \arg \max_k n_k$ be the class whose count is largest. Let $n_A$ and $n_B$ be the largest count and the second-largest count, respectively.
%Let $n_A$ and $n_B$ denote the number of classes $k_A$ and $k_B$ with the most and second most occurrences, respectively.
%Let $n_A$,$n_B$ denote the number of class $k_A$ and $k_B$ which appear the largest and second number of times, respectively.
If $k_A$ appeares much more often than any other class, then the prediction of MedRDF $g$ returns $k_A$. Otherwise, it abstains from making a prediction. 
As Cohen et al. \cite{cohen2019certified} declared, we use the hypothesis test from Hung \& Fithian \cite{hung2019rank} to calibrate the abstention threshold so as to bound by $\alpha$ the probability of returning an incorrect answer. The prediction of MedRDF $g$ satisfies the following guarantee:

\begin{table*}
	\centering
	\caption{Accuracy (\%) of different defense mechanism (rows) against white box adversarial attacks with maximum $L_\infty$ perturbation $\epsilon=8/255$ (columns) on \textbf{COVID-19 and DermaMNIST datasets with ResNet-18}. The original accuracy of each defense is described in the column ``Natural". GS: gaussian smoothing, MF: median filter. The number after attack method represents the number of iteration steps.}
	\centering  
	{
		{
			\begin{tabular}{c|c|c|c| cccc|  ccc| c}
				
				\toprule
				\multirow{2}*{Dataset}&\multirow{2}*{Method} & \multirow{2}*{Denoiser} &\multirow{2}*{\tabincell{c}{Natural}}  & \multicolumn{4}{c|} {I-FGSM}& \multicolumn{3}{c|} {PGD} & \multirow{2} *{C\&W}  \\ 
				\cmidrule{5-11}
				&&&&1&2&5&7&7&20&100& \\
				\midrule
				%			\specialrule{0.01em}{0.7pt}{0.7pt}
				\multirow{14}*{COVID-19}&\multirow{3}*{ResNet-18}& None & 94.4&5.6&0.0&0.0&0.0&0.0&0.0&0.0&0.0\\
				&& GS& 94.4&5.6&1.4&0.0&0.0&0.4&0.0&0.0&0.0\\
				&& MF& 94.4&35.2&23.8&1.2&0.2&1.8&0.0&0.0&0.0\\
				\cmidrule{2-12}
				%           \specialrule{0.01em}{0.7pt}{0.7pt}
				&\multirow{3}*{\tabincell{c}{MedRDF \\ with gaussian noise}}& None & 28.8& 40.3& 51.1&36.4&35.5&30.8&28.8&27.7&20.8 \\
				&& GS& 94.8&39.4&72.8&48.6&45.4&73.8&55.0&47.6&55.4 \\
				&& MF& 94.2&68.4&87.6&75.2&73.8&89.4

				&82.2&78.2&82.4\\
				\cmidrule{2-12}
				%           \specialrule{0.01em}{0.7pt}{0.7pt}
				&\multirow{3}*{\tabincell{c}{MedRDF \\ with s.p. noise}}& None & 65.4&28.4&44.6&24.8&23.8&39.4

				&26.2&21.2&26.0\\
				&& GS& \textbf{95.6}&42.6&79.6&53.2&49.0&82.2&64.2&54.0&64.0\\
				&& MF&
				91.2&\textbf{84.6}&\textbf{89.8}&\textbf{86.6}&\textbf{86.2}&\textbf{93.2}

				&\textbf{90.2}&\textbf{89.6}&\textbf{90.4}\\
				\cmidrule{2-12}
				&\multirow{3}*{\tabincell{c}{MedRDF\\ with poisson noise}}& None & 32.4& 28.4& 29.8&28.4& 28.2&29.4&28.6&28.6&28.8 \\
				&& GS&94.4& 45.2& 75.6&54.6&50.8&78.0& 62.8&55.6&62.6 \\
				&& MF&92.2&71.2& 86.6&77.2&75.6&89.0& 81.0&79.0&81.4
				\\
				\midrule
				\multirow{14}*{DermaMNIST}&\multirow{3}*{ResNet-18}& None & \textbf{74.1}&1.7&4.2&0.0&0.0&0.1&0.0&0.0&0.0\\
				&& GS& 71.5&12.5&25.2&2.5&1.5&16.2&1.1&0.4&0.9\\
				&& MF& 72.5&55.3&53.2&31.3&22.3&38.2&15.1&4.0&16.7\\
				\cmidrule{2-12}
				%           \specialrule{0.01em}{0.7pt}{0.7pt}
				&\multirow{3}*{\tabincell{c}{MedRDF \\ with gaussian noise}}& None & 44.6&5.4&26.7&7.0&5.1&26.0&12.3&10.1&16.9\\
				&& GS& 67.9&44.6&59.1&49.7&47.9&60.9&53.3&49.8&54.8\\
				&& MF& 70.9&61.1&\textbf{66.8}&\textbf{63.5}&62.7&\textbf{67.4}&64.8&63.3&65.8\\
				\cmidrule{2-12}
				%           \specialrule{0.01em}{0.7pt}{0.7pt}
				&\multirow{3}*{\tabincell{c}{MedRDF \\ with s.p. noise}}& None & 68.5&15.6&45.9&15.0&9.4&44.6&17.1&8.3&24.6\\
				&& GS& 68.4&48.1&60.5&52.2&50.9&62.4&56.0&53.5&57.4\\
				&& MF& 69.0&\textbf{62.1}&66.5&\textbf{63.5}&\textbf{63.1}&67.3&\textbf{65.1}&\textbf{64.1}&\textbf{66.0}\\
				\cmidrule{2-12}
				&\multirow{3}*{\tabincell{c}{MedRDF\\ with poisson noise}}& None & 58.9&13.4&38.0&16.7&11.6&37.0&18.9&11.4&26.5
				\\
				&& GS& 68.3&47.5&59.1&51.9&51.0&60.8&55.0
				&52.0&57.4
				\\
				&& MF& 69.4&60.0&65.0&62.0&61.5&66.3&63.9&62.9&65.3\\
				\bottomrule
			\end{tabular}
		}
	}
	\label{table:defence_resnet18}
\end{table*}

\begin{table*}[t]
	\centering
	\caption{Accuracy (\%) of different defense mechanism (rows) against white box adversarial attacks with maximum $L_\infty$ perturbation $\epsilon=8/255$ (columns) on \textbf{COVID-19 and DermaMNIST dataset  with ResNet-50}. The original accuracy of each defense is described in the column ``Natural". GS: gaussian smoothing, MF: median filter. The number after attack method represents the number of iteration steps.}
	\centering  
	{
		{
			\begin{tabular}{c|c|c|c| cccc|  ccc| c}
				
				\toprule
			\multirow{2}*{Dataset}&\multirow{2}*{Method} & \multirow{2}*{Denoiser} &\multirow{2}*{\tabincell{c}{Natural}}  & \multicolumn{4}{c|} {I-FGSM}& \multicolumn{3}{c|} {PGD} & \multirow{2} *{C\&W}  \\ 
				\cmidrule{5-11}
				&&&&1&2&5&7&7&20&100& \\
				\midrule
				%			\specialrule{0.01em}{0.7pt}{0.7pt}
				\multirow{14}*{COVID-19}&\multirow{3}*{ResNet-50}& None & 92.6& 62.8& 7.0&0.0&0.0&0.0&0.0&0.0 &0.0 \\
				&& GS& 92.6& 62.8& 7.0& 0.2&0.0&0.0&0.0&0.0&0.0 \\
				&& MF& 92.6& 70.2& 17.8&1.0&0.2&0.4&0.0&0.0&0.0 \\
				\cmidrule{2-12}
				%           \specialrule{0.01em}{0.7pt}{0.7pt}
				&\multirow{3}*{\tabincell{c}{MedRDF \\ with gaussian noise}}& None & 82.6& 68.2& 77.6&73.2&72.2&78.2&76.4&73.4&73.6 \\
				&& GS& 91.4& 74.6& 84.6&73.0&67.8&87.0&80.6&71.0&70.2 \\
				&& MF& \textbf{93.6}& 88.2& 91.8&88.8&87.2&\textbf{93.0}&91.4&89.4&89.0 \\
				\cmidrule{2-12}
				%           \specialrule{0.01em}{0.7pt}{0.7pt}
				&\multirow{3}*{\tabincell{c}{MedRDF \\ with s.p. noise}}& None & 89.8& 78.4& 86.6&82.6&81.6&87.6&85.4&82.2&83.2 \\
				&& GS& 90.4& 83.8& 88.8&85.6&84.4&88.8&88.4&87.6&86.8 \\
				&& MF& 91.6& \textbf{91.2}& \textbf{91.8}&\textbf{91.8}&\textbf{91.2}&91.6&\textbf{91.4}&\textbf{91.2}&\textbf{91.2} \\
				\cmidrule{2-12}
				%           \specialrule{0.01em}{0.7pt}{0.7pt}
			    &\multirow{3}*{\tabincell{c}{ MedRDF \\ with poisson noise }}&  None & 77.0& 66.6&75.4&69.8& 68.4&75.4&73.0&69.2&72.4 \\
			    && GS& 88.0& 80.8& 83.4&82.0& 80.4&83.6&83.0& 81.6&83.2 \\
			    &&MF& 87.8&83.8&86.4&85.0&84.0&88.2&87.4&85.6&87.0\\
				\midrule
				\multirow{14}*{DermaMNIST}&\multirow{3}*{ResNet-50}& None & 73.0&8.9&18.8&1.1&1.0&9.3&1.0&0.9&0.1\\
				&&GS&71.5&4.9&11.9&0.8&0.4&5.8&0.4&0.4&0.1\\
				&&MF&\textbf{73.2}&35.3&40.4&11.5&4.6&26.4&4.7&1.5&4.9\\
				\cmidrule{2-12}
				%           \specialrule{0.01em}{0.7pt}{0.7pt}
				&\multirow{3}*{\tabincell{c}{MedRDF \\with gaussian noise}}& None & 69.9&13.5&34.7&6.6&4.2&32.9&7.1&5.4&6.6\\
				&& GS&67.3&29.8&52.7&27.7&25.8&54.4&31.6&30.0&32.4\\
				&& MF&72.0&54.6&73.1&55.0&53.0&75.9&60.0&57.9&62.0\\
				\cmidrule{2-12}
				%           \specialrule{0.01em}{0.7pt}{0.7pt}
				&\multirow{3}*{\tabincell{c}{MedRDF \\ with s.p. noise}}& None & 71.3&12.7&32.4&6.2&4.3&31.0&8.4&6.2&8.1\\
				&& GS& 65.3&30.3&53.0&30.0&28.6&56.0&35.0&33.1&35.8\\
				&&MF&68.7&\textbf{60.8}&\textbf{75.5}&\textbf{62.8}&61.3&\textbf{78.5}&67.3&66.4&69.1\\
				\cmidrule{2-12}
					%           \specialrule{0.01em}{0.7pt}{0.7pt}
				&\multirow{3}*{\tabincell{c}{MedRDF \\ with poisson noise}}& None &70.0&21.4&38.5&19.1&17.3&39.0&21.2&19.3&21.8\\
				&& GS& 64.5&33.8&56.2&35.9&34.7&59.1&40.9
				&38.5&41.0\\
				&& MF&70.3&60.5&74.0&62.5&\textbf{62.7}&77.3&\textbf{68.1}&\textbf{67.2}&\textbf{70.0}\\
				\bottomrule
			\end{tabular}
		}
	}
	\label{table:defence_resnet}
\end{table*}

\begin{table*}[t]
	\centering
	\caption{Accuracy (\%) of different defense mechanism (rows) against white box adversarial attacks with maximum $L_\infty$ perturbation $\epsilon=8/255$ (columns) on \textbf{COVID-19 and DermaMNIST dataset with AG-Sononet-16}. The original accuracy of each defense is described in the column ``Natural". GS: gaussian smoothing, MF: median filter. The number after attack method represents the number of iteration steps.} 
	\centering  
	{
		{
			\begin{tabular}{c|c|c|c| cccc|  ccc| c}
				
				\toprule
				\multirow{2}*{Dataset} &\multirow{2}*{Method} & \multirow{2}*{Denoiser} &\multirow{2}*{Natural}  & \multicolumn{4}{c|} {I-FGSM}& \multicolumn{3}{c|} {PGD} & \multirow{2} *{C\&W}  \\ 
				\cmidrule{5-11}
				&&&&1&2&5&7&7&20&100& \\
				\midrule
				\multirow{14}*{COVID-19}&\multirow{3}*{AG-Sononet-16}& None & \textbf{93.4}& 29.2& 12.0&0.8&0.2&0.2&0.0&0.0&0.0 \\
				&& GS& 90.4& 41.8& 10.0&0.0&0.0&16.2&0.0&0.0&0.0 \\
				&& MF& 92.0 &75.6& 39.6& 17.6&10.6&26.8&3.0&0.0&0.0 \\
				\cmidrule{2-12}
				&\multirow{3}*{\tabincell{c}{MedRDF \\with gaussian noise}}& None & 28.2& 28.2& 28.2&28.2&28.2&28.2&28.2&28.2&28.2 \\
				&& GS& 75.0& 44.8& 64.6&53.0&50.4&66.4&60.2&49.6&52.4 \\
				&& MF& 88.6& 85.6& 87.4&85.6&84.2&87.2&86.2&83.6&83.8 \\
				\cmidrule{2-12}
				&\multirow{3}*{\tabincell{c}{MedRDF \\with s.p.  noise}}& None& 28.2& 28.2& 28.2&28.2&28.2&28.2&28.2&28.2&28.2 \\
				&& GS& 92.0& 65.2& 79.2&64.2&58.8&83.0&73.2&60.6 & 60.0 \\
				&& MF& 87.0& \textbf{88.0}& 87.0&\textbf{86.4}&83.8&\textbf{89.2}&\textbf{88.8}&\textbf{88.6}&\textbf{88.6} \\
				\cmidrule{2-12}
				&\multirow{3}*{\tabincell{c}{MedRDF \\ with poisson noise}}& None &28.2&28.2&28.2&28.2&28.2&28.2&28.2&28.2&28.2\\
				&& GS&88.8&84.2&88.0&85.6&\textbf{85.4}&88.6&87.8&86.4&87.2\\
				&&MF&90.0&86.4&\textbf{90.0}&86.0&83.4&88.7&88.4&86.8&87.3\\
				\midrule
				\multirow{14}*{DermaMNIST}&\multirow{3}*{ AG-Sononet-16}& None & \textbf{70.6}&38.7&31.4&11.5&5.5&17.7&2.7&0.1&3.5\\
				&& GS&68.2&42.0&38.4&24.3&18.8&31.9&19.3&6.8&19.3\\
				&& MF& 71.4&55.1&49.4&35.7&30.7&36.4&24.0&9.6&25.7\\
				\cmidrule{2-12}
				%           \specialrule{0.01em}{0.7pt}{0.7pt}
				&\multirow{3}*{\tabincell{c}{MedRDF \\ with gaussian noise}}& &42.2&25.5&36.4&29.4&27.1&37.9&31.6&27.3&32.4\\
				&&GS&65.6&57.4&62.4&60.3&59.8&64.4&63.4&62.6&63.5\\
				&& MF& 70.5&63.2&68.8&66.2&65.9&\textbf{70.3}&\textbf{69.5}&\textbf{68.7}&\textbf{69.5}\\
				\cmidrule{2-12}
				%           \specialrule{0.01em}{0.7pt}{0.7pt}
				&\multirow{3}*{\tabincell{c}{MedRDF \\ with s.p. noise}}& None &66.4&45.1&56.8&46.6&43.4&59.5&53.6&47.5&55.0\\
				&& GS& 66.3&58.5&64.3&62.7&61.7&66.4&65.5&64.9&65.5\\
				&& MF&68.3&\textbf{66.0}&\textbf{67.4}&66.1&\textbf{66.3}&68.9&68.5&68.2&68.8
				\\
				\cmidrule{2-12}
				%           \specialrule{0.01em}{0.7pt}{0.7pt}
				&\multirow{3}*{\tabincell{c}{MedRDF \\ with poisson noise}}& None & 61.9&45.4&54.5&47.6&45.1&56.6&52.1&48.3&53.4\\
				&& GS& 68.5&58.9&65.7&62.9&62.8&66.7&66.3
				&65.7&66.2\\
				&& MF&70.3&64.3&67.2&\textbf{66.5}&\textbf{66.3}&69.2&68.8&68.0&69.0\\
				\bottomrule
			\end{tabular}
		}
	}
	\label{table:defence_sononet}
\end{table*}

\begin{prop}
	With probability at least $1 - \alpha $  over the randomness in the prediction of MedRDF, the probability that the prediction of MedRDF returns a class other than $\hat{k_A}$ is at most $\alpha$, i.e.,
\end{prop}
\begin{equation}
\mathbb{P}(g(\bm{x}) \ne \hat{k_A}) \le \alpha.
\label{equ:proposition}
\end{equation} 
We use the p-value of the two-sided hypothesis test that $n_A$ is drawn from binomial distribution Binom($n_A+n_B, 1/2$) to verify whether Equation~(\ref{equ:proposition}) holds. If the p-value is less than $\alpha$, then return $k_A$. Else, abstain. i.e., we can adopt two-sided hypothesis test with binomial distribution ($\text{Binom}$) to justify the randomness in the prediction of MedRDF:
\begin{equation}
\text{Binom}(n_A,n_A+ n_B,1/2) \le \alpha.
\end{equation}
The proof is as follows:
\begin{pro}
MedRDF returns a class other than $\hat{k_A}$ if and only if (1) $k_A \neq \hat{k_A}$ and (2) MedRDF does not abstain.
We have:
\begin{equation}
\begin{split}
&\mathbb{P}(g(\bm{x})\neq \hat{k_A}) = \mathbb{P}(k_A \neq \hat{k_A}, \text{MedRDF does not abstain})  \\
 &= \mathbb{P}(k_A \neq \hat{k_A}) \mathbb{P}(\text{MedRDF does not abstain} | k_A \neq \hat{k_A}) \\
 &\leq \mathbb{P}(\text{MedRDF does not abstain}|k_A \neq \hat{k_A}).
\end{split}
\end{equation}
\end{pro}
Recall that MedRDF does not abstain if and only if the p-value of the two-sided hypothesis test that $n_A$ is drawn from Binom$(n_A+n_B, 1/2)$ is less than $\alpha$. Theorem 1 in Hung \& Fithian \cite{hung2019rank} proves that the conditional probability that this event occurs given that $k_A \neq \hat{k_A}$ is exactly $\alpha$. That is,
\begin{equation}
	\mathbb{P}(\text{MedRDF does not abstain}|k_A \neq \hat{k_A}) = \alpha.
\end{equation}
Therefore, we have:
\begin{equation}
\mathbb{P}(g(\bm{x}) \ne \hat{k_A}) \le \alpha.
\end{equation}
When $\alpha$ is small, MedRDF abstains frequently but rarely returns the wrong class. When $\alpha$ is large, MedRDF usually makes a prediction, but may often return the wrong class. 
$\alpha = 0.001$ and $n=1e^4$ have been adopted in our framework. The complete prediction procedure of MedRDF $g$ is described in Algorithm \ref{alg:framework}.

\subsection{Robust Metric}
In medical diagnostic tasks, in addition to the diagnostic results output by the model, we also hope to obtain the confidence score of the results, so as to better guide the doctor's follow-up work, such as adopting the result or re-evaluating this case. 
Therefore, in this subsection, in order to provide doctors with intuitive and effective indicator, we define a Robust Metric (RM) based on MedRDF. The formulation of RM is presented as follows:
\begin{equation}
\text{RM}=  \frac{K* (n_A-n_B )}{n},
\label{equ:rm}
\end{equation}
where $n_A$ and $n_B$ denote the number of classes $k_A$ and $k_B$ with the most and second most occurrences of $g$, respectively. $K$ is number of diagnosis categories. 
Setting the threshold of RM, when the RM output by MedRDF is greater than the threshold, the doctor can accept this diagnostic result. Otherwise, doctor should consider re-evaluating this result. The effectiveness of RM is analyzed as follows:

From Equation (\ref{equ:rm}) we can obtain:
\begin{equation}
(P_{k_A})_{\min}=(\frac{n_A}{n})_{\min} =\frac{1}{K}+\frac{K-1}{K^2}\text{RM}.
\end{equation}
Then for different classification tasks, doctors can set different thresholds to make the probability of output labels reach their expectations.
Take the 3-class diagnostic task as an example, we set a threshold of RM with $1$ for indicating the diagnostic result to be robust or not, that is to say, $k_A$ should have at least $5/9$ probability for robust evaluation.
For 7-class diagnostic task, when setting the threshold of RM as $3$, the probability of class $k_A$ is at least 0.51.

%For example, 2-class dignosis task at least should have $3/4$ probability on $k_A$ and 3-class dignosis task at least should have $5/9$ probability for robust evaluation.
%\resizebox{0.98\textwidth}{0.16\textheight}

\section{Experiments}
\label{experiment}
In this section, we first introduce the experimental settings including the common isotropic noises and adversarial attack methods we used in this study. 
Second, we choose the best noise boundry $\sigma$ and the number $n$ of the copies for our experiments by ablation study.
Then, we conduct a set of experiments to evaluate the robustness of our MedRDF under different adversarial attacks. Furthermore, we confirm the necessity and effectiveness of our RM indictor and visually present the robust diagnostic results for different cases. Finally, we have shown more comparable results on our MedRDF with other augmentation strategies and other defense methods.
\subsection{Experimental Settings}

\subsubsection{Common Isotropic Noise}
We evaluate the robustness of MedRDF under gaussian noise, salt-and-pepper (s.p.) noise and poisson noise, and utilize gaussian smoothing (GS) and median filter (MF) as denoisers in experiments. 
\subsubsection{Adversarial Attack}
The adversarial examples are crafted by the most challenging ``white-box" attacks (i.e., I-FGSM\cite{kurakin2016adversarial}, PGD\cite{madry2018towards}, and C\&W \cite{carlini2017towards}) and ``black-box" attacks (i.e., SPSA~\cite{uesato2018adversarial} and RayS~\cite{chen2020rays}).
The ``white-box" attacks are under maximum $L_\infty$ perturbation $\epsilon = 8/255$.

\subsection{Ablation Study}
\subsubsection{The level of common noise and adversarial perturbation} We first explore the influence of common noise $\sigma$ and adversarial perturbation $\epsilon$ on original and robust accuracy. As shown in TABLE~\ref{table:dnoise}, since we set $\epsilon=8/255$ for adversarial attack, we choose $\sigma=0.1$ for the boundry of the common noise for its high accuracy. 

\subsubsection{The number of the copies}
In this part, we explore the influence of different number of copies to the final robustness of MedRDF. The defense accuracy and test time of MedRDF on different number $n$ of copies are recorded in TABLE~\ref{table:n}. As shown in TABLE~\ref{table:n}, both on COVID-19 and DermaMNIST datasets, although the natural accuracy on $n=1e^5$ is higher than it on $n=1e^4$, the test time on each image when $n=1e^5$ is much longer than it on $n=1e^4$, which is not conducive to the clinical application of the MedRDF. For example, on COVID-19 dataset, the natural accuracy is 91.4\% on $n=1e^5$, which is little higher than 91.2\% on $n=1e^4$. However, its test time on each image is 87.6s, which is not easily tolerated when compared with the test time 3.8s on $n=1e^4$.
Besides, in terms of defense accuracy, it can be seen that $n=1e^4$ has a greater impact on the final defense accuracy on MedRDF, compared with that on $n=1e^3$ and $n=1e^5$. For instance,  with $n=1e^4$, the accuracy on DermaMNIST attacked by C\&W is 66.0\%, which is higher than the accuracy (65.6\% and 65.9\%) of $n=1e^3$ and $n=1e^5$, respectively.
In summary, we choose the number $n=1e^4$ in our experiments.
%\textbf{Robust performance under attacker on pre-trained models $h'_{\bm{\theta}}$.}  

\subsection{Robustness Evaluation and Analysis.}
\subsubsection{Quantitative Results} 
In this part, we present the quantitative results of base model and MedRDF under different white-box attacks and black-box attacks.

\textbf{White box attack.} The accuracy of original models (i.e., $h_{\bm \theta}$ = ResNet-18, ResNet-50, and AG-Sononet-16) and our MedRDF (i.e., $g_{\bm \theta}$ based on ResNet-18, ResNet-50, and AG-Sononet-16) are recorded in TABLE~\ref{table:defence_resnet18}, TABLE \ref{table:defence_resnet} and TABLE \ref{table:defence_sononet}, respectively.
As shown in TABLE~\ref{table:defence_resnet18}, the original model ResNet-18 is vulnerable to adversarial attacks both on COVID-19 and DermaMNIST diagnostic tasks (e.g., its accuracy drops to 0.0\% under C\&W attack on COVID-19 and DermaMNIST). The other result we can observe from TABLE~\ref{table:defence_resnet18} is that, MedRDF markedly improves the robustness of original model in all attack settings (e.g., when under PGD-7 attack, the accuracy of MedRDF with gaussian noise and MF denoiser is 67.4\% while the original pre-trained model is 0.1\% on DermaMNIST).
The same result can also be found in TABLE~\ref{table:defence_resnet} on ResNet-50 and TABLE \ref{table:defence_sononet} on AG-Sononet-16. As shown in TABLE \ref{table:defence_resnet}, MedRDF even maintains a better performance on natural accuracy (i.e., the natural accuracy on COVID-19 of MedRDF with gaussian noise and MF denoiser is 93.6\%, while the original ResNet-50 is 92.6\%).
As shown in TABLE~\ref{table:defence_sononet}, one can observe that the classification results of two datasets have been significantly improved after using MedRDF. For instance, on DermaMNIST dataset, the defense accuracy of MedRDF with gaussian noise and median filter is 70.3\% when attacked by PGD-7, which is much better than the accuracy of base model AG-Sononet-16 (i.e., 17.7\%), and even is comparable with that without any attack (i.e., natural accuracy 70.6\%). 
These results indicate the effectiveness of our framework to convert non-robust models to robust ones. 

%The same results can also be found in Table \ref{table:defence_sononet} on AG-Sononet-16.
\begin{table}
	\centering
	\caption{Accuracy (\%) of original model ResNet-50 and AG-Sononet-16 on COVID-19 under different settings. The common noise $\sigma=0.1$,  GS: gaussian smoothing, MF: median filter.} 
	\centering  
	{
		\setlength{\tabcolsep}{1.5mm}{
			\begin{tabu}{c|c|c|c|c|c|c|c}
				\toprule
				\multirow{2}*{Method} &\multirow{2}*{Natural}& \multicolumn{3}{c|}{Gaussian Noise}& \multicolumn{3}{c}{Salt-and-Pepper Noise}\\
				\cmidrule{3-8}
				&&None& GS&MF& None& GS&MF\\
				\cmidrule{1-8}
				ResNet-50 & 92.6 & 81.8 & 91.2& 93.4& 90.0 & 90.6& 91.2 \\
				AG-Sononet-16 & 93.4 & \textbf{28.2} & 87.4& 88.0& \textbf{28.2} &85.0& 85.6\\
				\bottomrule
			\end{tabu}
	}}
	\label{table:noise}
\end{table}

\begin{table}[t]
	\centering
	\caption{Accuracy (\%) of different defense mechanism (rows) against \textbf{black box adversarial attacks} on COVID-19 and DermaMNIST dataset with \textbf{ResNet-18}. GS: gaussian smoothing, MF: median filter, s.p. noise: salt-and-pepper noise.} 
	\centering  
	{
			\begin{tabular}{c|c|cc|cc}
				\toprule
				\multirow{2}*{Method} &\multirow{2}*{Denoiser}& \multicolumn{2}{c}{COVID-19}& \multicolumn{2}{c}{DermaMNIST}\\
				\cmidrule{3-6}
				&&SPSA& RayS&SPSA& RayS\\
				\midrule
				\multirow{3}{*}{ResNet-18} & None & 0.7 & 0.0& 0.0&0.0 \\
				& GS & 2.7 & 1,0& 0.9&3.7 \\
				& MF & 36.0 & 1.4& 15.8&6.9 \\
				\midrule
				
				\multirow{3}{*}{\tabincell{c}{MedRDF \\ with gaussian noise}} & None & 52.7 & 64.6& 63.2&
				70.7\\
				& GS & 86.7 & 85.6& 68.8&69.9\\
				& MF & 86.7 & 86.2& \textbf{71.4}&\textbf{72.7}\\
				\midrule
				
				\multirow{3}{*}{\tabincell{c}{MedRDF \\ with s.p. noise}} & None & 66.0 & 66.0& 56.5&71.5
				 \\
				& GS & 86.5 & 86.4& 68.7&70.2\\
				& MF & 86.7 & 86.6& 71.3&72.3\\
				\midrule
				
				\multirow{3}{*}{\tabincell{c}{MedRDF \\ with poisson noise}} & None & 60.5 & 60.6& 60.7&
				70.0\\
				& GS & 86.6& 86.6& 68.8&69.5 \\
				& MF & \textbf{86.7} & \textbf{86.6}& 70.8&71.7\\
				\bottomrule
			\end{tabular}
	}
	\label{table:black_resnet18}
\end{table}

\begin{table}[t]
	\centering
	\caption{Accuracy (\%) of different defense mechanism (rows) against \textbf{black box adversarial attacks} on COVID-19 and DermaMNIST dataset with \textbf{AG-Sononet-16}. GS: gaussian smoothing, MF: median filter, s.p. noise: salt-and-pepper noise.} 
	\centering  
	{
		\begin{tabular}{c|c|cc|cc}
			\toprule
			\multirow{2}*{Method} &\multirow{2}*{Denoiser}& \multicolumn{2}{c}{COVID-19}& \multicolumn{2}{c}{DermaMNIST}\\
			\cmidrule{3-6}
			&&SPSA& RayS&SPSA& RayS\\
			\midrule
			\multirow{3}{*}{AG-Sononet-16} & None & 9.3 & 0.0& 1.8&0.0 \\
			& GS & 1.3 & 0.1& 13.9&3.9\\
			& MF & 31.3 & 13.5& 20.1& 5.0  \\
			\midrule
			
			\multirow{3}{*}{\tabincell{c}{MedRDF \\ with gaussian noise}} & None & 32.5 & 32.2& 64.2& 69.4 \\
			& GS & 80.6 & 80.6& 66.6& 66.7  \\
			& MF & 82.2 & 82.0& 70.8& 71.0  \\
			\midrule
			
			\multirow{3}{*}{\tabincell{c}{MedRDF \\ with s.p. noise}} & None & 68.6 & 68.7& 62.1& 70.8 \\
			& GS & 84.0 & 84.0& 66.7& 66.8  \\
			& MF & 81.9 & 82.0& 71.4& 71.9  \\
			\midrule
			
			\multirow{3}{*}{\tabincell{c}{MedRDF \\ with poisson noise}} & None & 55.3 & 57.3& 63.9& 70.5 \\
			& GS & \textbf{84.6} & \textbf{84.0}& 67.2& 67.9  \\
			& MF & 82.6 & 82.0& \textbf{71.5}& \textbf{72.2}  \\
			\bottomrule
		\end{tabular}
	}
	\label{table:black_sononet}
\end{table}

\begin{figure*}[t]
	\centering
	\begin{subfigure}{1.0\textwidth}
		\centering
		\includegraphics[width=\textwidth]{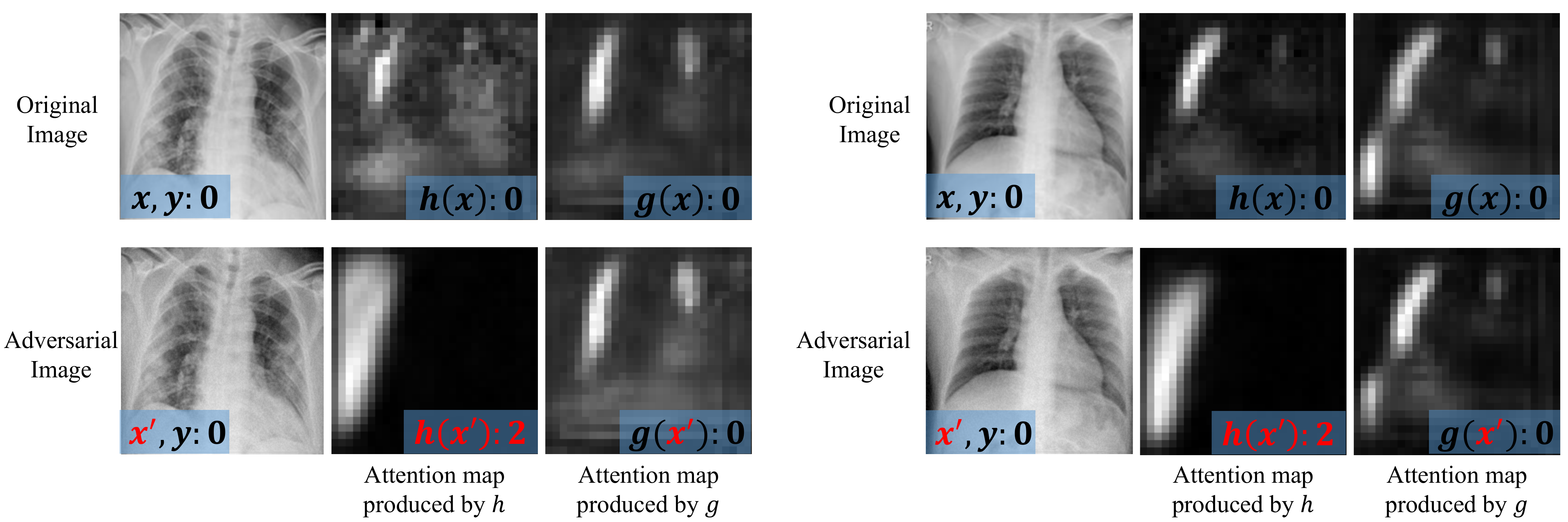}
		\caption{ COVID-19 Cases.}
	\end{subfigure}
	\begin{subfigure}{1.0\textwidth}
		\centering
		\includegraphics[width=\textwidth]{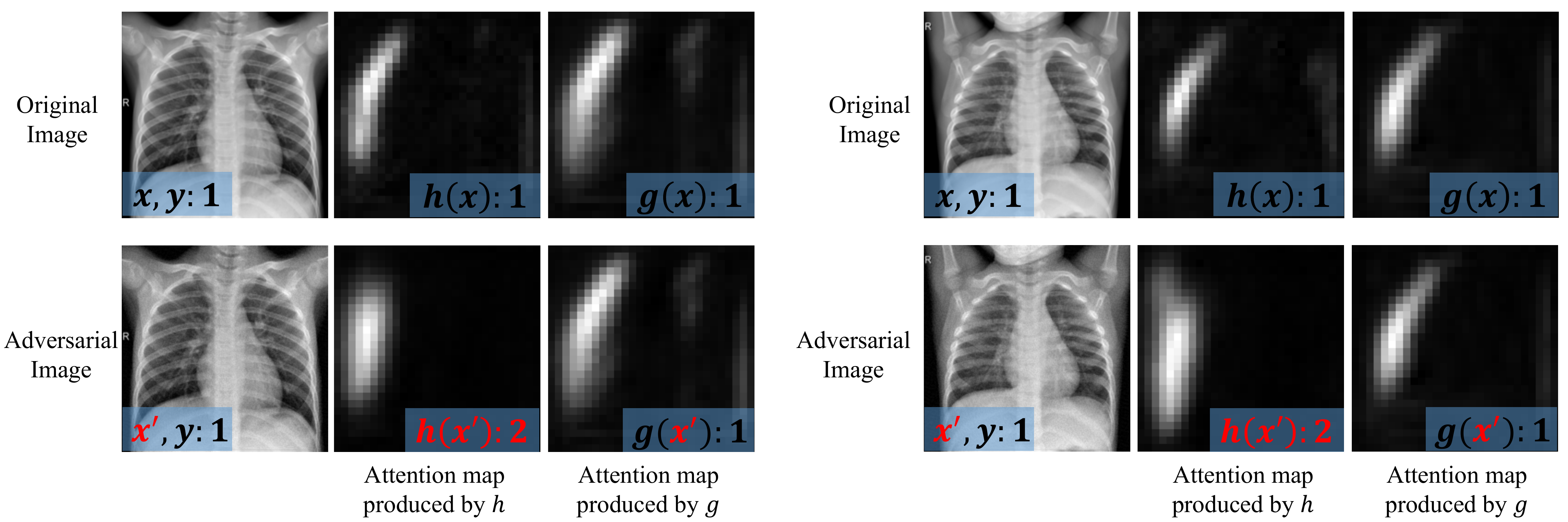}
		\caption{Normal Cases.}
	\end{subfigure}
	\begin{subfigure}{1.0\textwidth}
		\centering
		\includegraphics[width=\textwidth]{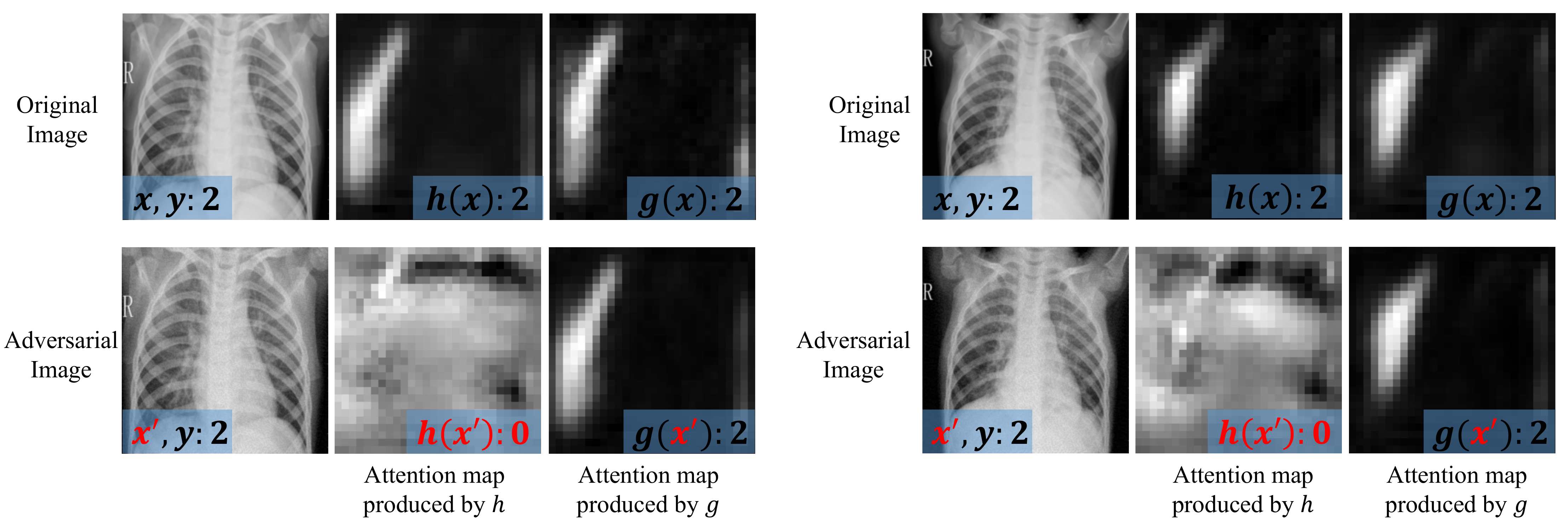}
		\caption{Viral Pneumonia Cases.}
	\end{subfigure}
	\caption{Attention maps of base model $h$ and MedRDF $g$ on original images and adversarial images. The first row of each subfigure contains original image and corresponding attentions maps of base model $h$ and MedRDF $g$, respectively. The second row of each subfigure contains adversarial image generated by PGD with 100 steps and corresponding attentions maps of base model $h$ and MedRDF $g$, respectively. Red denotes the adversarial images and wrong labels. Base model $h$ is AG-Sononet-16, MedRDF $g$ is based on AG-Sononet-16 architecture with salt-and-pepper noise and median filter.}
	\label{fig:adversarial_result_imdep}
\end{figure*}

\begin{figure*}[t]
	\centering
	\begin{subfigure}{0.3\textwidth}
		\centering
		\includegraphics[width=\textwidth]{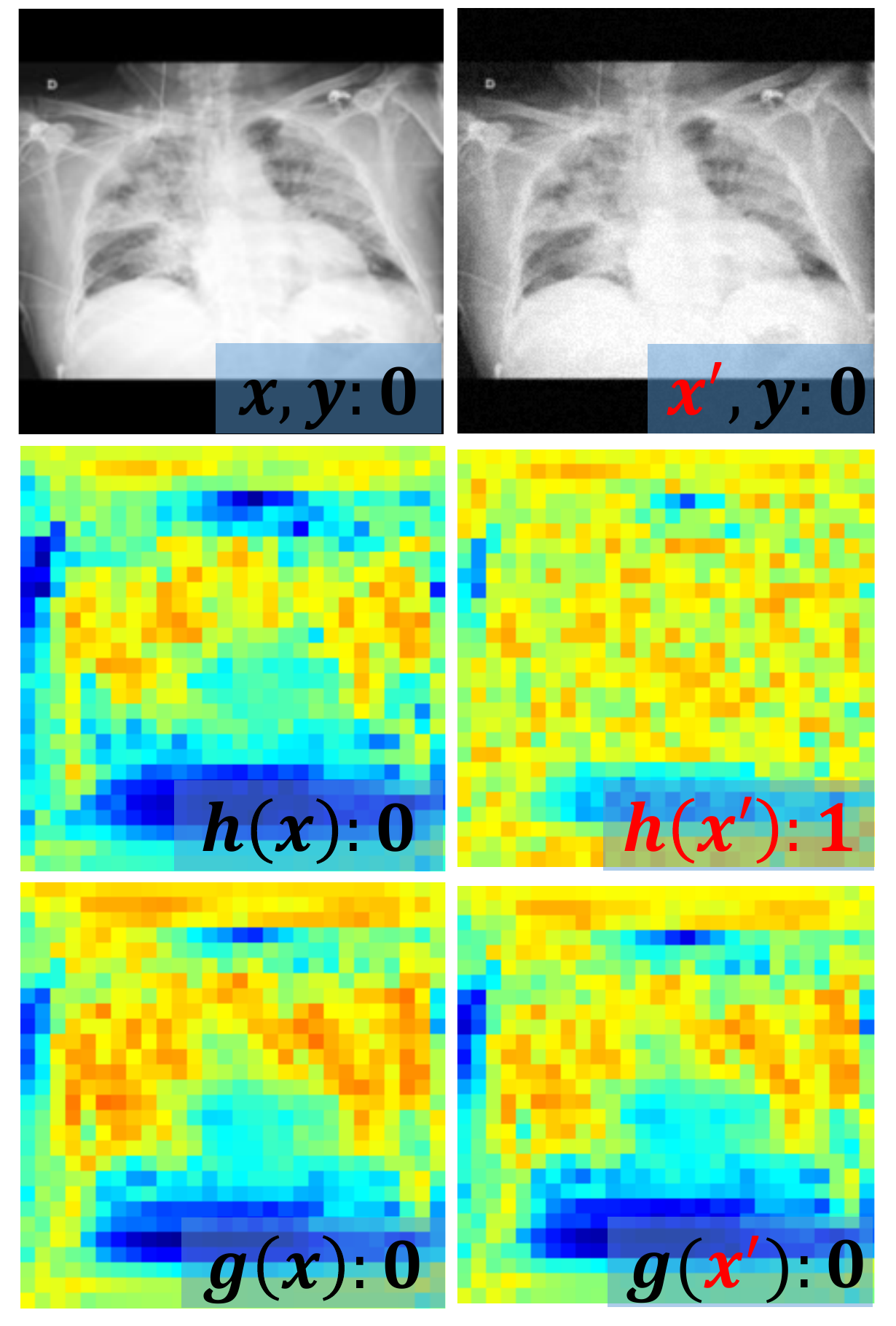}
		\caption{ COVID-19 Case.}
	\end{subfigure}
	\begin{subfigure}{0.3\textwidth}
		\centering
		\includegraphics[width=\textwidth]{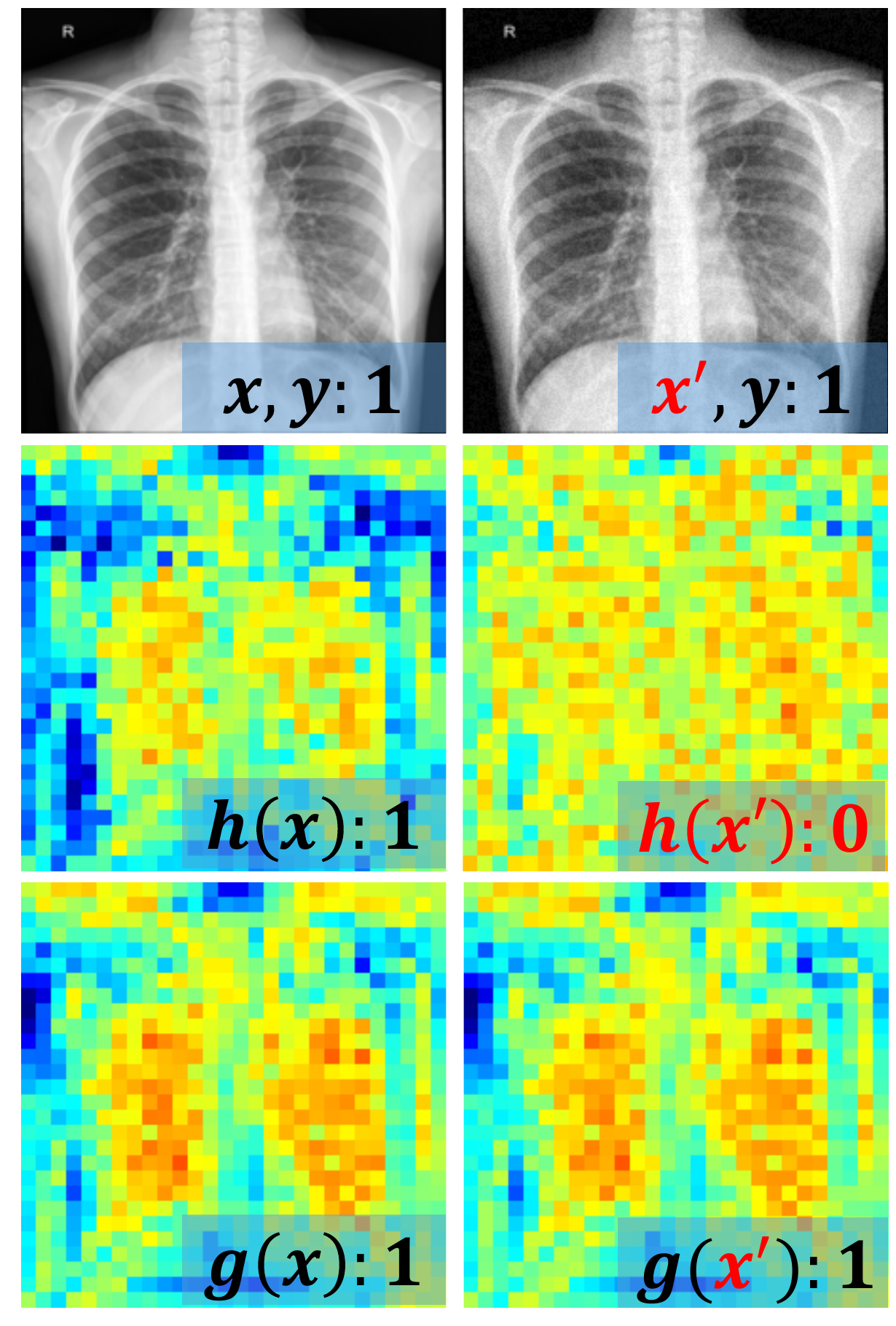}
		\caption{Normal Case.}
	\end{subfigure}
	\begin{subfigure}{0.3\textwidth}
		\centering
		\includegraphics[width=\textwidth]{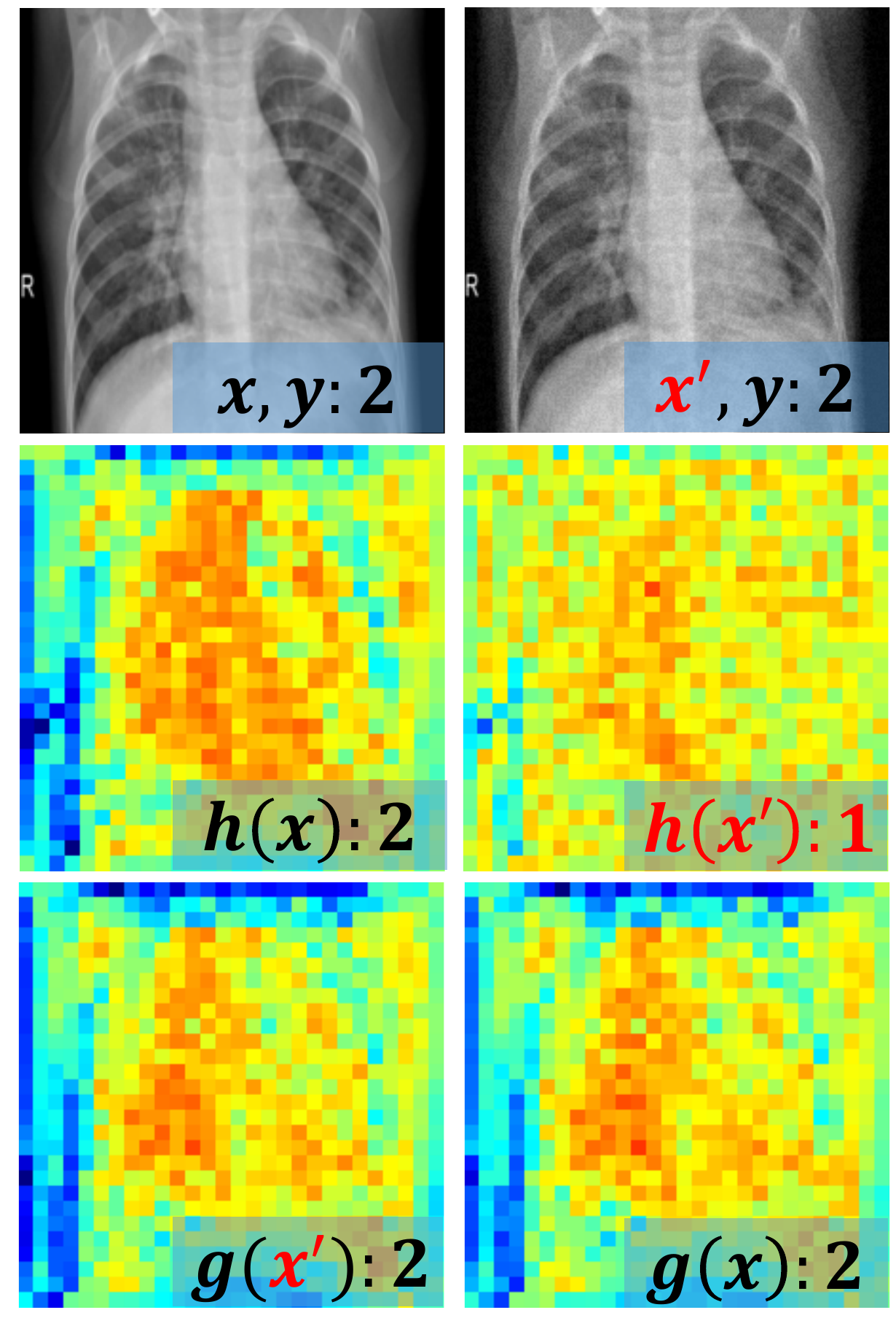}
		\caption{Viral Pneumonia Case.}
	\end{subfigure}
	\caption{Feature maps of base model $h$ and MedRDF $g$ on original images and adversarial images. The first row of each subfigure contains original image and its corresponding adversarial image. The second row of each subfigure shows the feature maps of original image and adversarial image on base model $h$, respectively. The third row of each subfigure shows the feature maps of original image and adversarial image on MedRDF $g$, respectively. The adversarial image is attacked by C\&W attack. The feature is at the second ``BasicBlock" layer of ResNet-18. MedRDF $g$ is based on ResNet-18 with salt-and-pepper noise and median filter.}
	\label{fig:heatmap}
\end{figure*}

\begin{figure*}[h]
	\centering
	\includegraphics[width=0.89\textwidth]{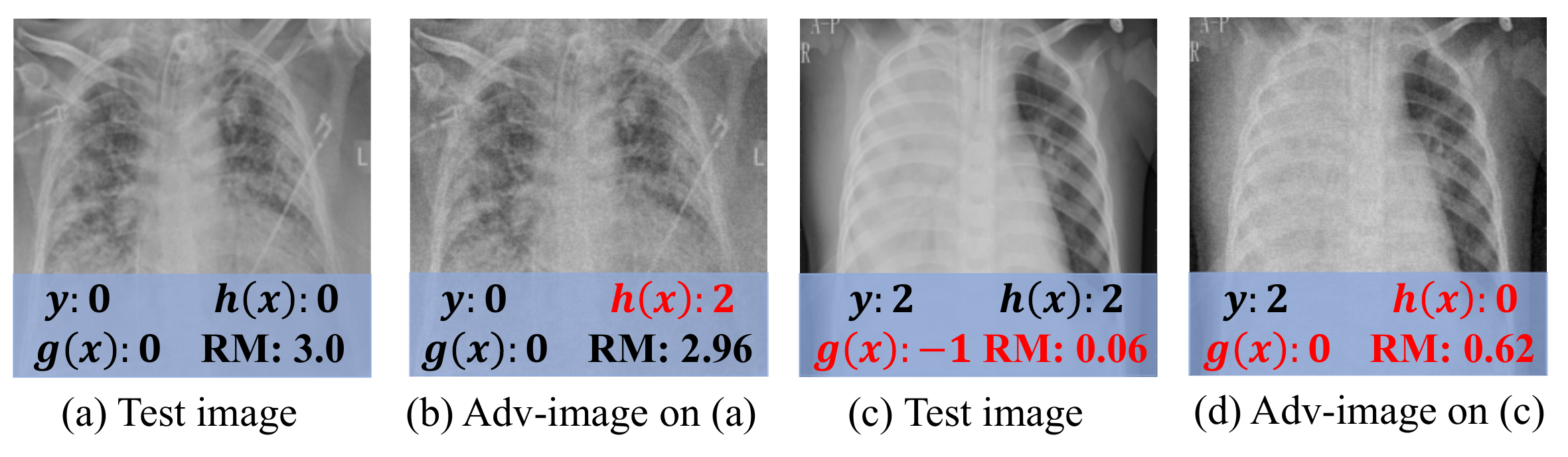}
	\caption{Robust Metric (RM) of several cases. (a) and (c) are selected from COVID-19 test set. (b) and (d) are generated by PGD with 100 steps on (a) and (c), respectively. The base model $h$ is ResNet-50, the MedRDF $g$ is based on ResNet-50 with gaussian noise and median filter.  Red represents the wrong label. Low RM indicates that (c) and (d) should be re-evaluated by professional doctor. } \label{fig:rm}
\end{figure*}

\begin{table}[h]
	\centering
	\caption{The accuracy (\%) produced by base classifiers (ResNet-50 and AG-Sononet-16) and MedRDF on original COVID-19 test set and adversarial test sets. Row ``Original" describes the original test set and rows ``PGD-100" and ``C\&W" represent the test sets attacked by PGD and C\&W. The abbreviation ``C." and ``R." represent Correctly classification and Robust evaluation ($\text{RM} \ge 1$), respectively. } 
	\centering  
	{
		\setlength{\tabcolsep}{0.25mm}
		{
			\begin{tabular}{c|c|c|c|c|c|c}
				\toprule
				\multirow{2}*{Network}&\multirow{2}*{Dataset}&\multirow{2}*{\tabincell{c}{Natural\\Acc $h_{\bm{\theta}}$}}  & \multicolumn{4}{c} {MedRDF $g$} \\
				\cmidrule{4-7}
				&&& C.\&R. & C.\&Not-R.& Not-C.\&Not-R.& Not-C.\&R. \\
				\cmidrule{1-7}
				\multirow{3}*{ResNet-50}&Original & 92.6 &90.8&0.6&7.2&1.4 \\
				&PGD-100 & 0.0 &90.2&1.2&7.8&0.8 \\
				&C\&W & 0.0 &90.4&1.0&7.8&0.8 \\
				\cmidrule{1-7}
				\multirow{3}*{\tabincell{c}{AG-Sononet\\-16}}&Original & 93.4 &84.6&2.4&11.0&2.0 \\
				&PGD-100 & 0.0 &87.4&1.8&9.2&1.6 \\
				&C\&W & 0.0 &87.0&2.2&9.2&1.6 \\
				\bottomrule
			\end{tabular}
	}}
	\label{table:rm}
\end{table}

\begin{table*}
	\centering
	\caption{Accuracy (\%) of different augmentation strategies (rows) against white box adversarial attacks with maximum $L_\infty$ perturbation $\epsilon=8/255$ (columns) on COVID-19 with ResNet-18. The original accuracy of each defense is described in the column ``Natural". The denoiser is MF after adding s.p. noise. The number after attack method represents the number of iteration steps.}
	\centering  
	{
		{
			\begin{tabular}{c|c|c| c|c|c|c}
				
				\toprule
				Method&Natural & I-FGSM-1& I-FGSM-7& PGD-20& PGD-100& C\&W \\ 
				\midrule
				%			\specialrule{0.01em}{0.7pt}{0.7pt}
				MedRDF with random rotating& 49.6& 28.6&29.2&28.8&28.7&29.0\\
				MedRDF with random resizing& 90.8& 51.4&57.0&62.2&55.6&60.0\\
				MedRDF with s.p. noise (ours)& \textbf{91.2} & \textbf{84.6} & \textbf{86.2} & \textbf{90.2} & \textbf{89.6} & \textbf{90.4} \\
				\bottomrule
			\end{tabular}
		}
	}
	\label{table:aug}
\end{table*}

\begin{table*}[t]
	\centering
	\caption{Accuracy (\%) and training time (hrs) of MedRDF compared with other defense methods under different adversarial attacks. The maximum $L_\infty $ adversarial perturbation is $\epsilon=8/255$, the numbers after the attack methods represent the number of iterative steps. MedRDF is based on salt-and-pepper noise and median filter denoiser. }
	\centering
	{
		\begin{tabular}{c|c|c|c|c|c|c|c|c}
			\toprule
			Dataset& Network&Method & Natural & I-FGSM-7 & PGD-20 & PGD-100 & C\&W & Training time (hrs) \\
			\midrule
			\multirow{12}*{\tabincell{c}{COVID-19}}&\multirow{6}*{\tabincell{c}{ResNet-50}}&Random R-P\cite{xie2018mitigating} & 53.0&49.8&51.2&50.6&51.0 & 0.66\\
		&&ComDefend\cite{jia2019comdefend} & 82.3&58.7&55.6&52.1&51.9 & 2.53\\			
			&&AT \cite{madry2018towards} & 90.8 & 65.6 & 65.0 & 61.4 & 62.0  & 4.17\\
			&&TRADES \cite{zhang2019theoretically}& 87.8 & 72.8 & 72.4 & 71.6 & 71.4 & 5.57\\
			&&MART \cite{wang2019improving}& 89.2  & 75.0 & 74.8 & 74.0 & 74.4 & 4.57\\
			&&\textbf{MedRDF} & \textbf{91.6} & \textbf{91.2} & \textbf{91.4} & \textbf{91.2} & \textbf{91.2} & \textbf{0.66}\\
			\cmidrule{2-9}
			
			&\multirow{6}*{\tabincell{c}{ResNet-18}}&
			Random R-P\cite{xie2018mitigating} & 66.2&56.2&56.2&54.8&56.4 &0.51\\
			&&ComDefend\cite{jia2019comdefend} &88.3&64.3&64.8&63.2&63.1 &1.30 \\			
			&&AT \cite{madry2018towards}& 95.2&67.6&67.8&65.0&67.0&1.50\\
			&&TRADES \cite{zhang2019theoretically}& 87.6&72.0&72.0&71.4&71.6&1.97\\
			&&MART \cite{wang2019improving}& 94.0&75.4&75.2&74.6&75.2&1.58\\
			&&\textbf{MedRDF} & \textbf{95.6} &\textbf{86.2} & \textbf{90.2} & \textbf{89.6} & \textbf{90.4} & \textbf{0.51}\\
			\midrule
			
			\multirow{12}*{\tabincell{c}{DermaMNIST}}&\multirow{6}*{\tabincell{c}{ResNet-50}}&
			Random R-P\cite{xie2018mitigating} &  {63.0}&{54.1}&{56.7}&{55.4}&{55.8} & {0.44}\\
			&&ComDefend\cite{jia2019comdefend} & {70.3}&{65.5}&{64.3}&{62.4}&{61.5} & {5.01} \\			
			&&AT \cite{madry2018towards} & {\textbf{72.5}}&{57.9}&{58.0}&{56.9}&{57.4}&{6.03}\\
			&&TRADES \cite{zhang2019theoretically}& {66.9}&{\textbf{66.8}}&{66.7}&{66.8}&{66.7}&{7.47}\\
			&&MART \cite{wang2019improving}& {70.6}&{64.0}&{64.1}&{63.5}&{62.9}&{5.14}\\
			&&\textbf{MedRDF}& {72.0}&{62.7}&{\textbf{68.1}}&{\textbf{67.2}}&{\textbf{70.0}}& {\textbf{0.44}}\\
			\cmidrule{2-9}
			
			&\multirow{6}*{\tabincell{c}{ResNet-18}}&
			Random R-P\cite{xie2018mitigating} & {50.3}&{40.4}&{41.4}&{40.1}&{41.9} &{0.44}\\
			&&ComDefend\cite{jia2019comdefend} & {68.3}&{62.4}&{61.8}&{55.2}&{55.1} &{1.50} \\			
			&&AT \cite{madry2018towards}& {\textbf{71.7}}&{56.5}&{56.7}&{55.1}&{55.6}&{1.54}\\
			&&TRADES \cite{zhang2019theoretically} &{68.7}&{62.8}&{64.5}&{\textbf{64.4}}&{63.9}&{2.03}\\
			&&MART \cite{wang2019improving}& {70.4}&{59.8}&{59.8}&{59.3}&{58.3}&{1.55}\\
			&&\textbf{MedRDF} & {69.0}&{\textbf{63.1}}&{\textbf{65.1}}&{64.1}&{\textbf{66.0}}&{\textbf{0.44}}
			\\
			\bottomrule
		\end{tabular}
	}
	\label{tab:comparison}
\end{table*}

Moreover, 
TABLE \ref{table:noise} records the accuracy of COVID-19 with original models ResNet-50 and AG-Sononet-16 under different noise settings. The result we can obtain from TABLE \ref{table:noise} is that, 
since original AG-Sononet-16 model is not robust to common noise (i.e., the natural accuracy is 28.2\% after adding noise without denoiser), the MedRDF without denoiser will lose its discrimination ability (i.e., the accuracy is 28.2\% under all attacks with random guess in TABLE \ref{table:defence_sononet}). This result has attracted our attention that the robustness of base model $h_{\bm{\theta}}$ under common noise will affect the final robustness of MedRDF under adversarial attack.

\textbf{Black box attack.} 
For SPSA attack, to estimate the gradients, we set the batch size as 128, the perturbation as $8/255$, and the learning rate as 0.01. We run SPSA attack for 100 iterations, and early-stop when we cause misclassification.
For RayS attack, we set the $L_\infty$ perturbation $\epsilon$ as $8/255$.
The accuracy of original models (i.e., $h_{\bm \theta}$ = ResNet-18 and AG-Sononet-16) and our MedRDF (i.e., $g_{\bm \theta}$ based on ResNet-18 and AG-Sononet-16) attacked by SPSA and RayS are recorded in TABLE~\ref{table:black_resnet18} and TABLE \ref{table:black_sononet}, respectively.
As shown in TABLE~\ref{table:black_resnet18}, the base model ResNet-18 with our MedRDF obtain better robustness in each attack. For instance, on COVID-19 dataset, MedRDF with poisson noise and median filter achieves 86.7\% accuracy when attacked by SPSA, which is much higher than that (0.7\%) on base ResNet-18.
Meanwhile, the same improvement can also be found on AG-Sononet-16 in TABLE \ref{table:black_sononet}. For example, on DermaMNIST, MedRDF with poisson noise and median filter achieves 72.2\% accuracy when attacked by RayS, while the accuracy of AG-Sononet-16 is 0.0\%.

\subsubsection{Visualization Results}
In this part, we illustrate the superiority of our proposed MedRDF by visualizing the changes in the internal feature of each model.

\textbf{The change of attention maps.} As shown in Fig.~\ref{fig:adversarial_result_imdep}, in each subfigure, the first column shows the original image and its corresponding adversarial image attacked by PGD-100 with their labels, respectively. The second column shows the attention maps and output labels of base model AG-Sononet-16 on original image and adversarial image, respectively. 
And the third column denotes the attention maps and output labels of MedRDF $g$ on original image and adversarial image, respectively. 
From Fig.~\ref{fig:adversarial_result_imdep} we can observe, the attention maps of base model on original image (i.e., $h(x)$) and adversarial image (i.e., $h(x')$) are extremely different. 
From this we can infer that, due to the changes of feature the base model focuses on, the base model can be easily fooled by adversarial example.
On the contrary, we notice that MedRDF do not significantly change the attention map of the original image and the adversarial example, which shows that MedRDF are more robust than the base model. It can effectively improve the robustness of the original model.

\textbf{The change of feature maps.} We have also shown the feature maps produced by base model ResNet-18 and MedRDF based on ResNet-18 in Fig.~\ref{fig:heatmap}. In each subfigure, the first row shows the original image  and adversarial image attacked by C\&W.  The second row shows the feature maps on original image and adversarial image at the second ``BasicBlock" layer of the base model ResNet-18, respectively. And the feature maps at the third row are produced by MedRDF. From the second rows at Fig.~\ref{fig:heatmap} (a)-(c), one can observe that the learned features of base model $h$ for the clean image focus on semantically informative regions (represented in red), while the features of the adversarial images are activated globally (without any specific focus). 
However, this problem can be effectively solved by MedRDF. From the third row of each subfigure, we can see that the feature map of the adversarial image generated by MedRDF is consistent with the clean image. These visualization results indicate that our MedRDF is not susceptible to adversarial perturbations, thus improving the robustness effectively.

\subsection{Performance of Robust Metric.} 
\subsubsection{Quantitative Results}

We report the accuracy produced by base models and MedRDF on original test set and adversarial test set in TABLE \ref{table:rm}, where the abbreviation ``C." and ``R." represent Correctly classification and Robust evaluation ($\text{RM} \ge 1$), respectively.
From TABLE \ref{table:rm} one can observe that, MedRDF can obtain robust and reliable accuracy both on original test dataset and adversarial test dataset (e.g., 90.8\% C.\&R. accuracy on original dataset and 90.4\% C.\&R. accuracy on adversarial dataset attacked by C\&W based on ResNet-50), even if the accuracy of original model on adversarial dataset drops to 0.0\%. 
In addition, we can find that for examples that were misclassified by MedRDF (i.e., Not-C in TABLE \ref{table:rm}),  most examples' RM are below the threshold (i.e., Not-C \& Not-R in TABLE \ref{table:rm}), which can effectively instruct the doctor to re-diagnose this case.
These results confirm the necessity and effectiveness of the RM indicator for medical diagnostic tasks.

\subsubsection{Visualization Results}
In order to illustrate the effectiveness of RM more intuitively, several cases are presented in Fig. \ref{fig:rm}, where the last two cases should be re-evaluated with doctor due to the low RM of the result. 

\subsection{Comparison with Other Methods}
\subsubsection{Comparison with Other Augmentation Strategies}
For each test image, MedRDF first creates a large number of noisy copies. To illustrate the effectiveness of this operator, we compare our operator of creating noisy copies with other augmentation strategies. Specifically, we use random resizing and random rotating to replace the noise in this experiment. The resizing range is $[200,224]$, the rotating angle is $[10,100]$. The experimental results can be found in TABLE~\ref{table:aug}. From TABLE~\ref{table:aug} we can obtain, compared with random rotating and resizing, our proposed MedRDF with noisy copies achieves best accuracy under all attacks.

\subsubsection{Comparison with Other Defense Methods.} 
To further verify the superior performance of our method, we compare MedRDF with other defense mechanisms in this section, including pre-processing based-defenses (i.e., Random R-P~\cite{xie2018mitigating}, ComDefend~\cite{jia2019comdefend}), and retraining-based defenses (i.e., adversarial training (AT)\cite{madry2018towards}, TRADES~\cite{zhang2019theoretically}, and MART~\cite{wang2019improving}). The accuracy and training time of each method can be found in TABLE~\ref{tab:comparison}. 
From TABLE~\ref{tab:comparison} we can obtain,
when the dataset is COVID-19, whether the base model is ResNet-18 or ResNet-50, MedRDF not only has the highest defense accuracy (e.g. the accuracy of MedRDF based ResNet-50 attacked by C\&W is 91.2\% while the Random R-P is 51.0\%), but its training time is much shorter than other retraining defense methods (e.g., the training time of MedRDF based ResNet-18 is 0.51 hrs while the TRADES is 1.97 hrs).
For DermaMNIST, MedRDF still maintains the best defense accuracy under many attacks (e.g., PGD-20, PGD-100, and C\&W on ResNet-50). Compared with Random R-P, the pre-processing defense method, MedRDF has better defense accuracy on medical images. Besides, compared with retraining methods, MedRDF which is employed in the inference phase can greatly reduce the training time and training burden.
The above analyses confirm that our MedRDF is effective and suitable for defending against adversarial attack on medical diagnostic tasks.
%that MedRDF acting in the inference phase is faster than the defense methods of retraining the network (i.e., AT, TRADES, and MART), and can also obtain higher accuracy.
%which confirms that our MedRDF is effective and suitable for medical diagnostic tasks.

%We show more experimental results in the supplementary material, including 1) the influence of common noise  $\sigma$ and  adversarial perturbation $\epsilon$ on original and robust accuracy on Table S2, 2) the inference time of original models and MedRDF on Table S3, and 3) visualization of the attention maps of original model and MedRDF in Fig. S1. 
%4). The accuracy and training time of MedRDF compared with other retrained defense methods (i.e., AT\cite{madry2018towards}, TRADES~\cite{zhang2019theoretically}, and MART~\cite{wang2019improving}) on Table A4.

\section{Conclusion}
\label{conclusion}
We propose a Robust and Retrain-Less Diagnostic Framework for Medical pre-trained models against adversarial attack (i.e., MedRDF). MedRDF allows users to seamlessly convert the pre-trained non-robust medical diagnostic model into robust one in inference phase, which is very convenient for diagnostic services that are already deployed online.
Moreover, we also propose an effective Robustness Metric (RM) based on MedRDF, which gives the confidence score of the diagnostic result. Experimental results demonstrate a superior performance of MedRDF on COVID-19 and dermaMNIST datasets in both white-box and black-box adversarial settings. 
In the future, we plan to study the robustness of base medical models to common noise which plays an important role in our robust framework, as well as the trade-off between the natural accuracy and defense accuracy. In addition, we will extend our research to the field of medical image segmentation.

\bibliographystyle{IEEEtran}
\bibliography{reference}

\end{document}